\documentclass[11pt]{article}
\usepackage{xcolor}
\usepackage{amsmath}
\usepackage{amssymb}
\usepackage{mathrsfs}
\usepackage{graphicx}
\usepackage{hyperref}
\usepackage{bm}
\usepackage{ucs}
\usepackage[utf8x]{inputenc}
\usepackage{wrapfig}
\usepackage{color}
\usepackage{float}
\usepackage{epstopdf}

\usepackage{amsmath,amstext,amssymb,graphicx,hyperref}
\usepackage{verbatim}
\usepackage{color}
\usepackage{algorithm,algorithmic}

\usepackage{mathtools}%AMS extensions

\usepackage{epstopdf}
\usepackage{color}
\usepackage{graphicx}
\usepackage{float}
\usepackage[section]{placeins}

\usepackage{tikz}
\usetikzlibrary{arrows}

\def\bfX{{\bf X}}

\def\Gc{{\cal G}}

\def\vect#1{\mbox{\boldmath{$#1$}}}

\def\vx{\vec{\vect x}}
\def\vy{\vec{\vect y}}
\newcommand{\bx}{\vect x}
\newcommand{\by}{\vect y}

\def\mE{\mathbb{E}}

\def\mC{\mathbb{C}}

\def\wG{\widehat{G}}

\def\wg{\widehat{g}}

\newcommand{\eps}{\varepsilon}
\renewcommand{\epsilon}{\eps}
\renewcommand{\leq}{\leqslant}
\renewcommand{\geq}{\geqslant}

\def\bfX{{\bf X}}
\def\bfY{{\bf Y}}
\def\bfD{{\bf A}}

\begin{document}

%\title{High resolution imaging in complex media}
\title{Wave-informed dictionary learning for high-resolution imaging in complex media}
%\author{Miguel Moscoso, Alexei Novikov, George Papanicolaou and Chrysoula Tsogka}

\author{Miguel Moscoso%
\thanks{Department of Mathematics, Universidad Carlos III de Madrid, Leganes, Madrid 28911, Spain, moscoso@math.uc3m.es}%
 \and 
 Alexei Novikov%
 \thanks{Department of Mathematics, Penn State University, University Park, PA 16802, novikov@psu.edu}%
 \and 
George Papanicolaou%
\thanks{Department of Mathematics, Stanford University, Stanford, CA 94305, papanicolaou@stanford.edu}%
\and 
Chrysoula Tsogka
\thanks{Department of Applied Mathematics, University of California, Merced, 5200 North Lake Road, Merced, CA 95343, ctsogka@ucmerced.edu}}

\maketitle
%%%%%%%%%%%%%%%%%%%%%%%%%%%%%
\begin{abstract}
We propose an approach for imaging in scattering media when large and diverse data sets are available. 
It has two steps. Using a dictionary learning algorithm the first step estimates the true Green's function vectors as columns in an unordered sensing matrix. 
The array data comes from many sparse sets of sources whose location and strength are not known to us. 
In the second step the columns of the estimated sensing matrix are ordered for imaging using Multi-Dimensional Scaling with connectivity information derived from cross correlations of its columns, as in time reversal. 
For these two steps to work together we need data from large arrays of receivers so the columns of the sensing matrix are incoherent for the first step, 
as well as from sub-arrays so that they are coherent enough to obtain connectivity needed in the second step. Through simulation experiments, 
we show that the proposed approach is able to provide images in complex media whose resolution is that of a homogeneous medium.
\end{abstract}
%\begin{keywords}
%array imaging, strongly scattering media, dictionary learning, multi-dimensional scaling
%\end{keywords}
%\section{Introduction}
%\label{sec:intro}
High-resolution imaging in complex media faces challenges due to wavefront distortion caused by scattering from inhomogeneities. 
In this paper we introduce a new approach for imaging in inhomogeneous, random media involving two basic components.
The first is a sparse dictionary learning algorithm in order to estimate Green's function vectors between focal or source points in the image window and receiver locations on the array.
The second is a Multi-Dimensional Scaling (MDS) algorithm to convert information about correlations of Green's function vectors into positions of the focal points in the image window. 
%of the Green's function vectors. 

 To accomplish the first step, we use a sparsity promoting modification of the Method of Optimal Directions (MOD)~\cite{Engan00} to learn an (unordered) dictionary of Green's function vectors that characterize the propagation of signals from a set of focal points, or sources in the image window, to the array. 
Here unordered means that we do not know which focal points are associated with the estimated column vectors of the dictionary.
In this step we assume that an abundance of %diverse 
 sensing measurements is available.  Specifically, we have access to measurements for multiple signals emanating from many sparse sets of sources, 
 but we do not have prior knowledge of their locations or amplitudes. 
 %\textcolor{red}{Given a suitable initialization,} 
 This dictionary learning method enables us to estimate Green's function vectors with high accuracy under the condition that these vectors are sufficiently incoherent, which means that their normalized inner product is sufficiently small.
 Given a configuration of sources or focal points in the image window, this implies that the receiver array must be large enough. We present this dictionary learning step in Section~\ref{sec:dl}.

The goal of the second step is to associate each Green's function vector with its corresponding focal point in the image window, which means that we want to find the correct order of the columns in the estimated matrix of Green's function vectors. 
%If we knew the medium's fluctuations, that is, if we knew each Green's function vector including its focal point, then we could migrate the estimated vectors to their focal points. This means that by taking inner products, or correlations, of the estimated Green's function vectors with the supposed known exact Green's function, only one would produce a normalized inner product close to one and this would identify its focal point.
%However, since these fluctuations are random, we do not have any prior knowledge of them. 
We could back propagate these vectors into the image window using a reference homogeneous medium. This is the Kirchhoff's migration approach 
that only works well when the fluctuations are weak \cite{Borcea05}. 
%The problem of finding the source location of the estimated Green's function vectors can be solved using 
For a given  set of Green's function vectors we could also try to estimate the position of their focal points using source localization algorithms, commonly used 
in wireless communications~\cite{Aspnes07,Stansfield47,Yeredor11,Wu19}. However, these algorithms use information based on distances between sources and receivers and are therefore very sensitive to noise. They are not suitable for imaging in media with strong fluctuations. 

Instead of estimating the distance between a focal point and a receiver, we can obtain a much more accurate estimate of the distance between two nearby focal points. 
The correlation of the estimated Green's function vectors gives such an estimate but\textcolor{red}{,} of course\textcolor{red}{,} we do not know where their focal points are located in the image window.
%Surprisingly the correlation provides a more accurate estimate in strongly scattering media due to super-resolution in time reversal, as demonstrated experimentally in~\cite{Fink92,Fink92b} and theoretically in \cite{Fink00,Borcea02}.
%\cite{Fink92,Lerosey04,Vellekoop07} 
By cross correlating each Green's function vector with all the others we can identify its nearest neighbors. This is a key observation that allows us to generate a proxy distance between column or Green's function vectors by counting the smallest number of neighborhoods that connect them.  This provides a connectivity-based proxy distance between all pairs of column vectors that we can use with the Multidimensional Scaling (MDS) algorithm \cite{Borg} for identifying Green's function vectors with their focal points up to a rotation, translation and scaling.  The resulting relative configuration of points can be spatially fixed with a few (two or three) known reference points in the two dimensional image window. The use of a proxy metric based on connectivity is done by the MDS-MAP algorithm~\cite{Shang03,Oh10}.  
%We use  a version of MDS to accomplish this identification of estimated Green's function vectors with focal points in the image window. 
Constructing the connectivity-based proxy distance using cross-correlations is described in Section~\ref{sec:grid}.

\section{Imaging problem setup}
\label{sec:setup}
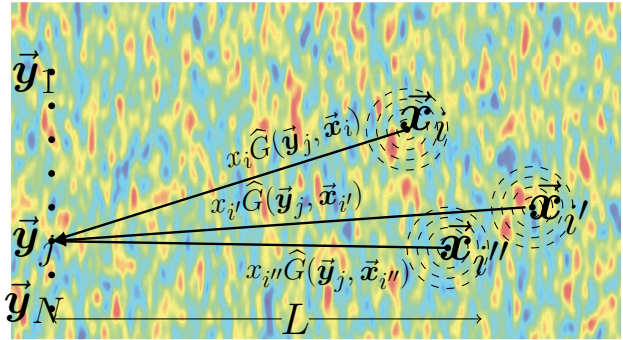
\begin{figure}[htbp]
    \centering
    \begin{tikzpicture}[scale=0.45, transform shape]
    \node[inner sep=0pt] (russell) at (5.8,4.1)
    {\includegraphics[width=18cm,height=10cm]{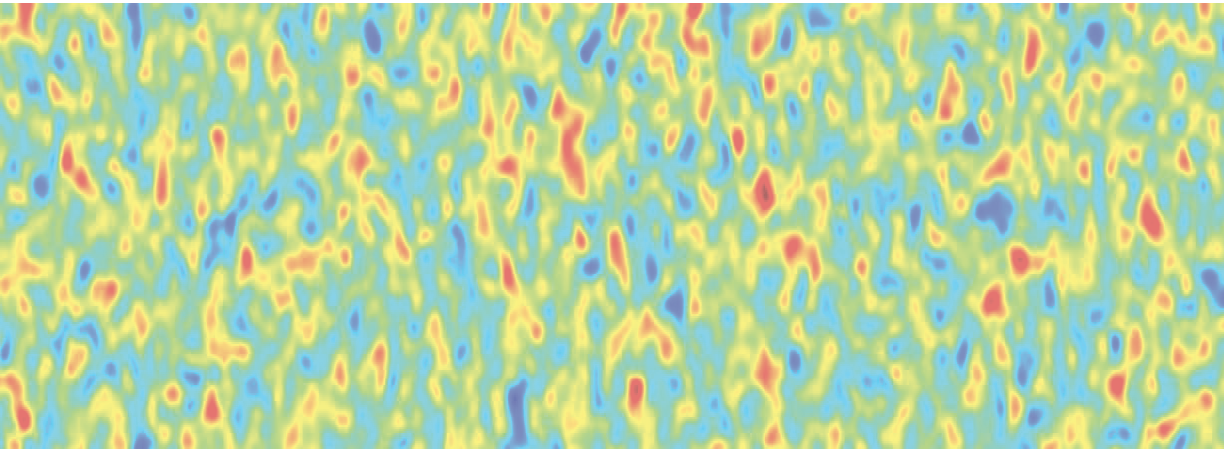}};
        % sensors
        \fill[fill=black] (-2, 0) circle [radius=0.1];
        \node[scale=1.5] at (-2.5, 0.2) {\Huge{$\vy_{N}$}};
        \fill[fill=black] (-2, 1) circle [radius=0.1];
        \fill[fill=black] (-2, 2) circle [radius=0.1];
        \node[scale=1.5] at (-2.5, 2) {\Huge{$\vy_{j}$}};
        \fill[fill=black] (-2, 3) circle [radius=0.1];
        \fill[fill=black] (-2, 4) circle [radius=0.1];
        \fill[fill=black] (-2, 5) circle [radius=0.1];
        \fill[fill=black] (-2, 6) circle [radius=0.1];
        \fill[fill=black] (-2, 7) circle [radius=0.1];
        \node[scale=1.5] at (-2.5, 7) {\Huge{$\vy_{1}$}};
        % horizontal axis
        \draw [<-] (-2, -0.3) -- (4.8, -0.3);
        \node[scale=1.5] at (5.2, -0.3) {\Huge{$L$}};
        \draw [->] (5.6, -0.3) -- (10.7, -0.3);
        % scatterers
        \fill[fill=black] (12.2, 3) circle [radius=0.1];
        \node[scale=1.8] at (13, 3) {\Huge{$\vx_{i'}$}};
        \fill[fill=black] (8.5, 5.3) circle [radius=0.1];
        \node[scale=1.8] at (9, 5.7) {\Huge{$\vx_{i}$}};
        \fill[fill=black] (9.7, 1.8) circle [radius=0.1];
        \node[scale=1.8] at (10.5, 1.8) {\Huge{$\vx_{i''}$}};
        % exciting fields
        \draw[dashed] (12.2, 3) circle [radius=0.3];
        \draw[dashed] (12.2, 3) circle [radius=0.6];
        \draw[dashed] (12.2, 3) circle [radius=0.9];
        \draw[dashed] (12.2, 3) circle [radius=1.2];
        \draw[dashed] (8.5, 5.3) circle [radius=0.3];
        \draw[dashed] (8.5, 5.3) circle [radius=0.6];
        \draw[dashed] (8.5, 5.3) circle [radius=0.9];
        \draw[dashed] (8.5, 5.3) circle [radius=1.2];
        \draw[dashed] (9.7, 1.8) circle [radius=0.3];
        \draw[dashed] (9.7, 1.8) circle [radius=0.6];
        \draw[dashed] (9.7, 1.8) circle [radius=0.9];
        \draw[dashed] (9.7, 1.8) circle [radius=1.2];
        % propagation to receiver
        \draw[-latex,line width=0.9pt] (12, 3) -- (-2, 2) node[pos=0.5, sloped, above]{\Huge{$x_{i'}\wG(\vy_j,\vx_{i'})$}};
        \draw[-latex,line width=0.9pt] (8.5, 5.3) -- (-2, 2) node[pos=0.3, sloped, above]{\Huge{$x_{i}\wG(\vy_j,\vx_{i})$}};
        \draw[-latex,line width=0.9pt] (9.7, 1.8) -- (-2, 2) node[pos=0.3, sloped, below]{\Huge{$x_{i''}\wG(\vy_j,\vx_{i''})$}};
    \end{tikzpicture}
    \caption{Schematic of the data collected on the array when three sources located at $\vx_i$, $\vx_{i'}$ and $\vx_{i''}$ simultaneously emit signals. It is the superposition of these signals that are recorded on the array of receivers located at points $\vy_j$, $j=1, \ldots, N$.}
    \label{fig:vectg}
\end{figure}

Suppose that an array of $N$ receivers records waves generated by sources located over a region of interest, called the image window. The receivers are located at points $\vy_j$, and the sources at unknown locations $\vx_i$. In Figure \ref{fig:vectg}, it is assumed that the array is 
%\sout{planar and rectangular} 
one-dimensional. The coordinates parallel to this array are the cross-range coordinates, and the ones orthogonal to it the range coordinates. %
The medium between the array and the unknown sources fluctuates randomly in space as illustrated in Figure \ref{fig:vectg}. The Green's function that characterizes  wave propagation in the random medium of a signal of frequency $\omega$ from a point $\vx$ %$\vect x$ 
to  a point $\vy$ % $\vect y$ 
satisfies the wave equation
\begin{equation}
\label{eq:wave equation}
\Delta\wG(\vx, \vy)+\kappa^2 \, n^2(\vx)\, \wG(\vx, \vy)=\delta(\vx - \vy),
\end{equation}
where  $\kappa=\omega/c_0$ is the wavenumber with $c_0$ a constant reference wave speed. The random index of refraction is $n(\vx)=c_0/c(\vx)$ with local wave speed $c(\vx)$. In a homogeneous medium,  $c(\vx)\equiv c_0$
for any location $\vx$ and, in this case, $\wG(\vx,\vy)=\wG_0(\vx,\vy)$, where
\begin{equation}\label{greenfunc}
\wG_0(\vx,\vy)=\frac{\exp( i \, \kappa \, |\vx-\vy|)}{4\pi|\vx-\vy|}\, .
%\quad \kappa=\frac{\omega}{c_0},
\end{equation}
%is the solution to \eqref{eq:wave equation}.
In random media, however, the wave speed $c(\vx)$ depends on the position $\vx$.
We consider a variable wave speed satisfying 
\begin{equation}\label{eq:random wave speed}
\frac{1}{c^2(\vx)}=\frac{1}{c_0^2}\bigg(1+\sigma\mu(\frac{\vx}{l})\bigg)\, ,
\end{equation}
where $l$ is the correlation length of the inhomogeneities that is characteristic of their size. In \eqref{eq:random wave speed}, 
$\sigma$ determines the strength of
the fluctuations around the constant speed $c_0$, and $\mu(\cdot)$ is a stationary random process with
zero mean and normalized autocorrelation function $R(|\vx_i-\vx_{i'}|)=\mE(\mu(\vx_i)\mu(\vx_{i'}))$, so $R(0)=1$.

We write the data received on the array of $N$ receivers in vector form with Green's function vector   
\begin{equation}\label{eq:GreenFuncVec}
\vect \wg(\vx)=[\wG(\vy_{1},\vx), \wG(\vy_{2},\vx),\ldots,
\wG(\vy_{N},\vx)]^{T}\,
\end{equation}
and we introduce the $N \times K$ sensing matrix
\begin{equation}\label{eq:sensingmatrix}
\vect{\Gc} =[\vect\wg(\vx_1)\,\cdots\,\vect\wg(\vx_K)]\, 
\end{equation}
defined on a grid $\{\vx_i\}_{i=1,\dots,K}$  spanning the image window. 
The sensing matrix in \eqref{eq:sensingmatrix} maps a distribution of sources in the image window to the (single frequency) data received on the array. 
The multi-frequency case is described in Section \ref{sec:experiments}.
For a given configuration of sources on the grid represented by the vector $\bx\in \mC^{K}$, the data
recorded on the array is given by
\begin{equation}
\label{eq:model}
\by = \vect{\Gc}\, \bx \, ,
\end{equation}
where $\by \in \mC^{N}$.
Here, $\bx$ is a vector whose $j$th component represents the complex amplitude of the source at location $\vx_j$ in the image window, ${j=1,\dots,K}$. 

Because the medium is random, the sensing matrix $\vect{\Gc}$ in \eqref{eq:model}  is not known. Our imaging problem is to estimate this matrix from a set of $M$ samples or observations $\{\by_i\}_{i=1,\dots,M}$, with 
$\by_i = \vect{\Gc}\, \bx_i$. The number of observations is large with respect to the dimension of the vectors $\bx_i$, i.e., $M \gg K$. Note that $\bx_i$ is also unknown but we assume that it is sparse, implying that the samples  $\by_i$ can be represented as a linear combination of a small number of columns of the unknown sensing matrix $\vect{\Gc}$. Since we do not know $\bx_i$, both the locations and the amplitudes of the sources are unknown. We assume that for the few sources that are active for every sample the modulus of their amplitude takes values in a bounded interval away from zero. % (i.e., $[0.6,0.8]$). 

\begin{figure}[htbp]
    \centering
%% left figure
% \makebox[0.4\textwidth]{
 { %\scalefont{4.5}
\begin{tikzpicture}[scale=0.2, transform shape]
\node[inner sep=0pt] (russell) at (24.8,4.1)
    {\includegraphics[width=.63\textwidth]{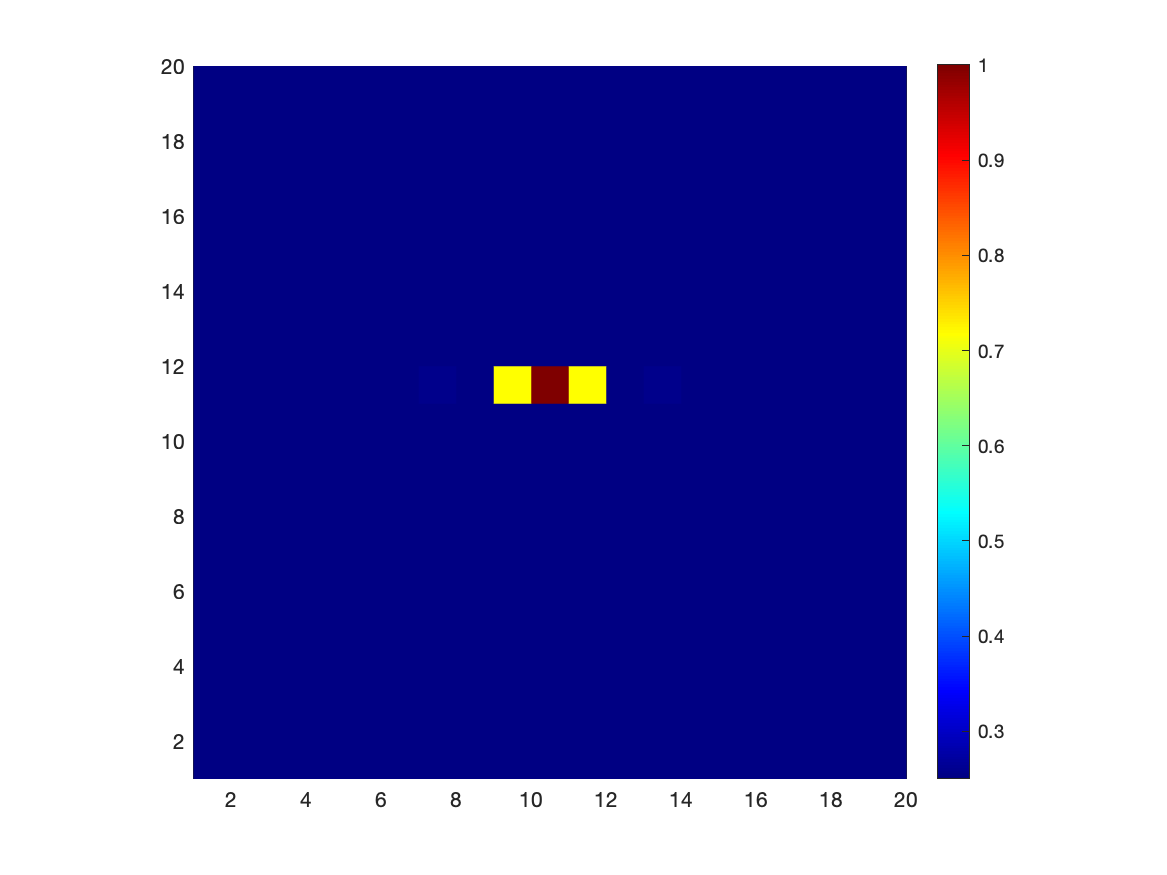}};
    \node[inner sep=0pt] (russell) at (19,-5)
    {\includegraphics[width=6cm,height=6cm]{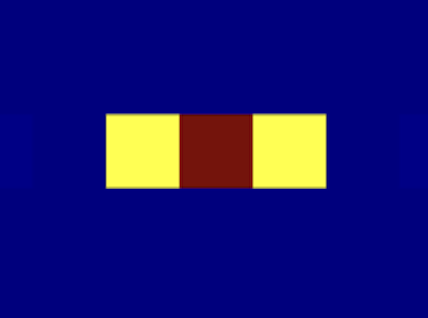}};

       \node at (6.5, 13.1) {(a)};
% vertical axis
        \draw [<-] (0, -5.1) -- (0, 3.);
        \node at (0, 4.0) {$a$}; %{\Large{$a$\Large}};
        \draw [->] (0, 5) -- (0, 13.1);
% % horizontal axis
        % sensors
         \fill[fill=black,opacity=1] (1.5, -5) circle [radius=0.1];
         \fill[fill=black] (1.5, -4) circle [radius=0.1];
         \fill[fill=black] (1.5, -3) circle [radius=0.1];
         \fill[fill=black] (1.5, -2) circle [radius=0.1];
         \fill[fill=black] (1.5, -1) circle [radius=0.1];
        \fill[fill=black] (1.5, 0) circle [radius=0.1];
        \fill[fill=black] (1.5, 1) circle [radius=0.1];
        \fill[fill=black] (1.5, 2) circle [radius=0.1];
        %\node at (1, 2) {\Large{$\vect x_s$}};
        \fill[fill=black] (1.5, 3) circle [radius=0.1];
        \fill[fill=black] (1.5, 4) circle [radius=0.1];
        \fill[fill=black] (1.5, 5) circle [radius=0.1];
        \fill[fill=black] (1.5, 6) circle [radius=0.1];
       % \node at (1, 6) {\Large{$\vect x_r$}};
       \fill[fill=black] (1.5, 7) circle [radius=0.1];
       \fill[fill=black] (1.5, 8) circle [radius=0.1];
       \fill[fill=black] (1.5, 9) circle [radius=0.1];
        \fill[fill=black] (1.5, 10) circle [radius=0.1];
        \fill[fill=black] (1.5, 11) circle [radius=0.1];
        \fill[fill=black] (1.5, 12) circle [radius=0.1];
        \fill[fill=black] (1.5, 13) circle [radius=0.1];
%        % echoes
   % arrows
        \draw[->,black, thick](2,-5) -- (24.5,4.25);
        \draw[->,black, thick](2,13) -- (24.5,4.8);    
        
       \draw[->,gray, thin](16,-2) -- (24.5,4);
       \draw[->,gray, thin](22,-2) -- (24.5,4);
\end{tikzpicture}

\vspace{-1.25cm}

%%right figure
\hspace{0.01cm}
\begin{tikzpicture}[scale=0.2, transform shape]
\node[inner sep=0pt] (Refocused spots) at (24.8,4.1)
    {\includegraphics[width=.63\textwidth]{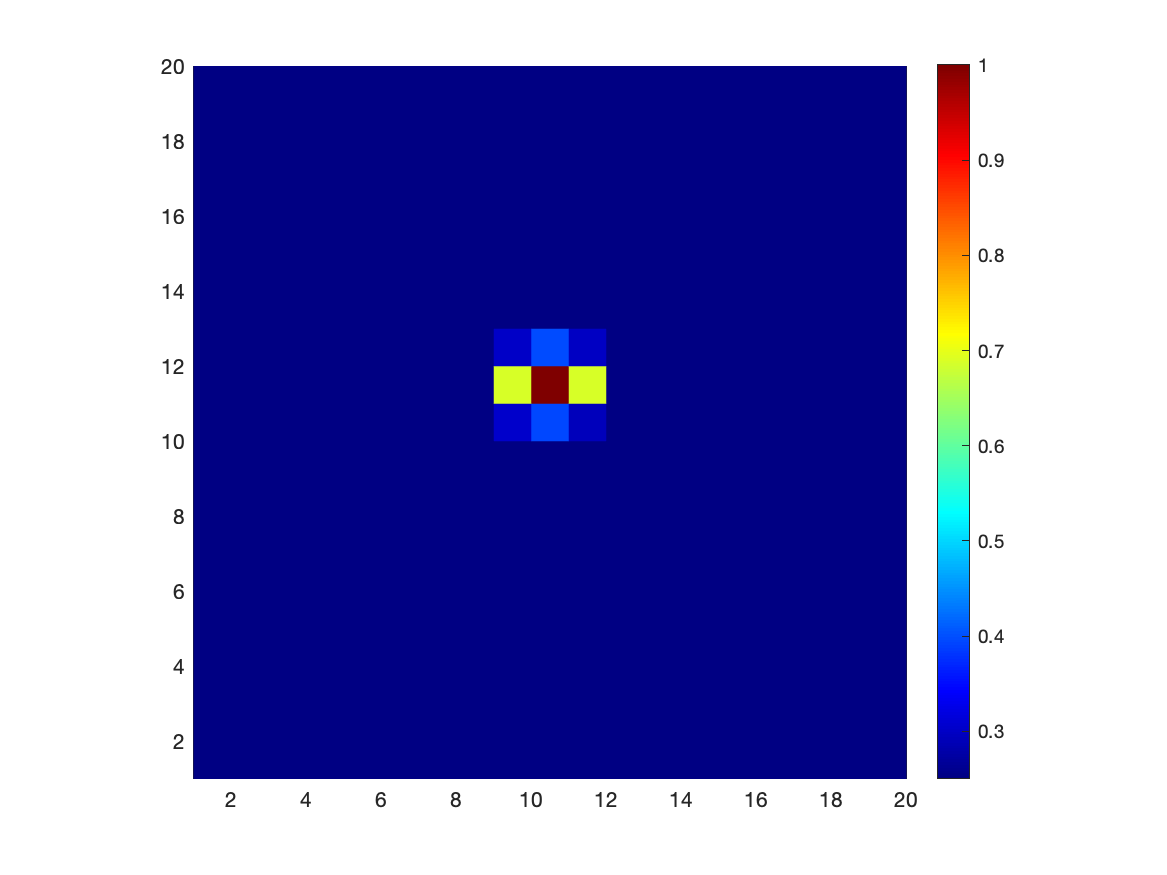}};
     %\node[inner sep=0pt] (russell) at (24.5,15)
     \node[inner sep=0pt] (russell) at (25,18.5)
    {\includegraphics[width=6cm,height=6cm]{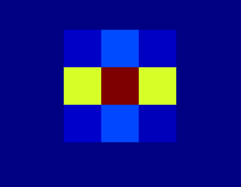}};
       \node[color=black] at (10, 19) {Zoom of the};
       \node[color=black] at (10, 17) {refocused spots};
       \node at (6.5, 13.1) {(b)};
% vertical axis
        \draw [<-] (0, -0.1) -- (0, 3.);
        \node at (0, 4.0) {$a/2$}; %{\Large{$a/2$\Large}};
        \draw [->] (0, 5) -- (0, 8.1);
 % horizontal axis
        \draw [<-] (1.7, -5.3) -- (12.8, -5.3);
        \node at (13.2, -5.3) {$L$};
        \draw [->] (13.6, -5.3) -- (25.7, -5.3);

        % sensors
         \fill[fill=black,opacity=0.3] (1.5, -5) circle [radius=0.1];
         \fill[fill=black,opacity=0.3] (1.5, -4) circle [radius=0.1];
         \fill[fill=black,opacity=0.3] (1.5, -3) circle [radius=0.1];
         \fill[fill=black,opacity=0.3] (1.5, -2) circle [radius=0.1];
         \fill[fill=black,opacity=0.3] (1.5, -1) circle [radius=0.1];
        \fill[fill=black] (1.5, 0) circle [radius=0.1];
        \fill[fill=black] (1.5, 1) circle [radius=0.1];
        \fill[fill=black] (1.5, 2) circle [radius=0.1];
        %\node at (1, 2) {\Large{$\vect x_s$}};
        \fill[fill=black] (1.5, 3) circle [radius=0.1];
        \fill[fill=black] (1.5, 4) circle [radius=0.1];
        \fill[fill=black] (1.5, 5) circle [radius=0.1];
        \fill[fill=black] (1.5, 6) circle [radius=0.1];
       % \node at (1, 6) {\Large{$\vect x_r$}};
       \fill[fill=black] (1.5, 7) circle [radius=0.1];
       \fill[fill=black] (1.5, 8) circle [radius=0.1];
       \fill[fill=black,opacity=0.3] (1.5, 9) circle [radius=0.1];
        \fill[fill=black,opacity=0.3] (1.5, 10) circle [radius=0.1];
        \fill[fill=black,opacity=0.3] (1.5, 11) circle [radius=0.1];
        \fill[fill=black,opacity=0.3] (1.5, 12) circle [radius=0.1];
        \fill[fill=black,opacity=0.3] (1.5, 13) circle [radius=0.1];
%        % echoes
     % arrows
        \draw[->,black, thick](2,0) -- (24.5,4.25);
        \draw[->,black, thick](2,8) -- (24.5,4.8);
        
        \draw[->,gray, thin](22.,15.5) -- (24.5,5);
        \draw[->,gray, thin](28.,15.5) -- (24.5,5);
        
               % \draw[thick, color=white] (24,13) grid (25,14);
\end{tikzpicture}
}
%}
\caption{Refocused spots with $(a)$ a large array used in the first step of the algorithm and $(b)$ a small array used in the second step. 
 }
    \label{fig:schematic}

\end{figure}
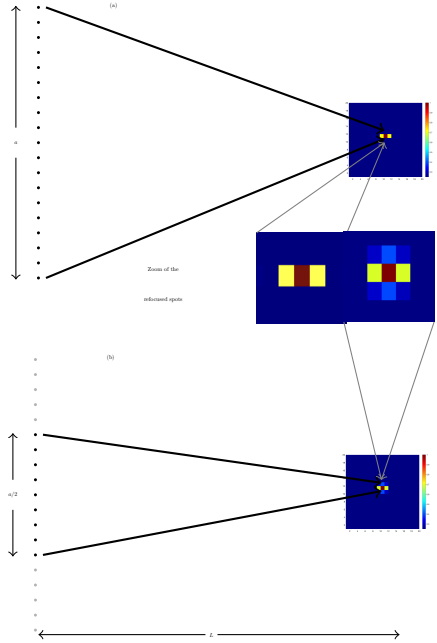
  
We note that for the imaging problem, the coherence between the columns of the sensing matrix  $\vect{\Gc}$ increases as the grid in the image window becomes finer. This can be challenging for the sparse dictionary learning algorithm, described in the next section,  as its convergence is guaranteed under incoherence or restricted isometry property assumptions \cite{Agarwal16}. Coherence is defined as the maximum of the normalized inner product between  different columns of the matrix, {\em i.e.}, 
\begin{equation}
\label{coh} 
\nu
=  \max_{\stackrel{i,j=1}{i \ne j}}^K   \frac{ \left|  \vect\wg(\vx_i)^* \vect\wg(\vx_j)  \right|}{|| \vect\wg(\vx_i) || ~||\vect\wg(\vx_j) ||}  .
\end{equation}
Given the size of the recording array and the bandwidth, in order to limit the coherence between the columns of $ \vect{\Gc}$ we assume that the grid in the image window is not finer than the support of the point spread function in a reference homogeneous medium; see Figure \ref{fig:schematic}-(a). This is essential for the first step of the approach, the dictionary learning step. 
The point spread function of an imaging system is the image one obtains when the signal from a single source is used as input.  

%The image computed by (\ref{eq:imag}) is the same as the one obtained by a time reversal experiment provided the Green's function of the medium is recovered with sufficient accuracy. %which is the case as suggested by our numerical simulations.  
The first step of our imaging algorithm uses dictionary learning and allows us to recover the columns of the sensing matrix 
%with high accuracy.  It is well known, however, that with sparse dictionary learning the columns of the dictionary are recovered 
up to a permutation. Although in most of the applications of dictionary learning the order of the columns is not an issue, it is essential in the imaging case.  That is because even though we recover the Green's function vectors, the imaging problem is still not solved as we do not know their correspondence with the grid points in the image window. To create an image we need to associate each column of the estimated matrix $ \hat{\vect{\Gc}}$ with the corresponding focal point in the image window. This is challenging since the propagation medium is unknown, so we cannot just back-propagate the recovered Green's function vectors, as it is done in time reversal.

In the second step of our approach, we deal with the {\em focal spot localization problem} where we associate each recovered Green's function vector $ \hat{\vect g}_i$ with its corresponding focal or source point in the image window.
The key idea is the use of cross-correlations between the columns of the estimated matrix $\hat{\vect{\Gc}}$ to identify the nearest neighbors of their focal or source points in the image window and then infer their associated focal points on the grid. Coherence becomes crucial in this step. 
%In this step we 
%need to increase the coherence between the estimated dictionary columns of $\vect{\Gc}_i$.
%This is because we 
%We use cross-correlationswe between the columns to identify the nearest neighbohrs of their focal or source points in the image window. 
We can do this by using for the cross-correlations a suitable subset of the elements of the estimated columns $ \hat{\vect g}_i$ corresponding to a fraction of the size of the recording array.
%This subset is chosen so that the point spread function of the imaging system is now bigger than the grid on which the sources are located, so that multiple grid points lie inside a point spread function. 
An example is illustrated in Figure \ref{fig:schematic}-(b) where a subarray of half the size is used.
%Thus, by using data from a sub-array as in  Figure \ref{fig:schematic}-(b), we can identify columns corresponding to neighbor points.  
The grid reconstruction problem is then solved using MDS with a proxy distance, as in sensor network localization problems \cite{Oh10}. No Euclidian distance information is known but a proxy distance is obtained from connectivity information at limited range that is recovered from cross-correlations. 

The cross-correlations formed can be interpreted as time reversal experiments. That is, the signals recorded on the array are re-emitted in the same medium and given the time-reversibility of the wave equation, those signals will focus at the location from which the original signal was emitted. Our connectivity reconstruction relies on this fundamental property of the wave equation and therefore is robust to the complexity of the medium. Our numerical simulations confirm that this is the case. %The limitation is absorption which breaks the time reversibility.   

\section{Dictionary learning}
\label{sec:dl}
In this Section, we discuss an algorithm aimed at learning from the gathered data a dictionary $\bfD \in \mC^{N\times K}$ that represents the normalized sensing matrix (\ref{eq:sensingmatrix}). We assume that the recorded signals $\by_i \in \mC^{N}$, $i=1,\dots,M$, the data, come from only a few sources and can therefore be represented as a linear combination of a small number of columns in the dictionary $\bfD$ we want to determine. 
This means that $\by_i = \bfD \, \bx_i$, where $\bx_i \in \mC^{K}$ are sparse vectors %with $\|\bx_i\|_0 \ll K$
that represent unknown collections of sources firing at the same time. For imaging applications, we can assume that the columns of $\bfD$ have unit lengths.
We also assume that we know the dimension $K$ (with usually $K\gg N$), which is the number of points in the image window and therefore specifies the resolution of the image. 
An estimate of $K$ can be based on the resolution of the imaging setup expected in a homogeneous medium. In a random medium\textcolor{red}{,} resolution in time reversal, but not in imaging, will improve  \cite{Fink00,Borcea02} and therefore $K$ could be larger. We assume here that $K$ is chosen based on a homogeneous medium.
%\textcolor{blue}{Because the dictionary $\bfD$ is usually overcomplete, meaning that $N<K$, there are infinitely many configurations of sources $\bx_i$ compatible with the same $\by_i$. This is where the sparsity constraint comes into play.}

To find the  dictionary $\bfD$ and the layout of sources, we define the matrix $\bfX=[\bx_1,\dots,\bx_M] \in \mC^{K\times M}$ and the data matrix  
$\bfY=[\by_1,\dots,\by_M] \in \mC^{N\times M}$,
and solve the problem 
\begin{equation}
%\label{dl_l0}
\label{dl_l1}
\begin{array}{rrclcl}
\displaystyle \min_{\bfD, \bfX} & \multicolumn{3}{l}{\| \bfD \bfX - \bfY\|_F^2 }\\
\textrm{s.t.} & \|\bx_i\|_{0}  \leq s, \,i=1,\dots,M, \\
%&\|\bd_k\|_2 =1,  k=1,\dots,K  \\
\end{array}
\end{equation}
where $ \|\cdot\|_0$ counts the number of non-zero elements and $s$ is the expected sparsity level.  
The decomposition $\bfY= \bfD \bfX $ is unique up to permutations of the columns of $\bfD$ and rows of $\bfX$ provided that the data $\bfY$ is rich enough \cite{Aharon06}. 

How much data do we need, that is, how big should $M$ be? As already noted, in imaging we usually have $K\gg N$. If the sparsity $s$ is fixed independent of $N$ then the condition
\begin{equation}
\label{eq:Mcond}
M >  K \log K
\end{equation}
is sufficient \cite{Agarwal17,Novikov23} for a suitable probabilistic model of ${\bf A,X,Y}$.

%one needs $O(N)$ measurements if the sparsity level is $s \leq \sqrt{N}$ under a suitable random model~\cite{Novikov23}.

%For example,  one needs $O(M)$ measurements if the sparsity level is $s \leq \sqrt{K}$ under a suitable random model~\cite{Novikov23}. \textcolor{blue}{Shouldn't be this $O(N)$ measurements if the sparsity level is $s \leq \sqrt{N}$?}

Problem (\ref{dl_l1}) is non-convex, as the constraint is not convex and both $\bfD$ and $\bfX$ are unknown. However, its solution can be found efficiently
by means of an alternating optimization procedure that uses the $\ell_1$-norm instead of the sparsity count, provided the initialization is close enough to the true solution and the columns of $\bf X$ are sparse enough \cite{Agarwal16}.  
Specifically, if $\bfD$ is known, then $\bfX$ in~\eqref{dl_l1} can be obtained as an $\ell_1$-norm minimization problem
\begin{equation}
\label{l1}
\displaystyle \min \|{\bfX}\|_1 \,\, \textrm{ subject to } \bfD \bfX = \bfY\, ,
\end{equation}
that can be easily solved by several different algorithms \cite{Davis97,Osborne00,Daubechies04,Beck09}.
%by using several algorithms such as Orthogonal Matching Pursuit (OMP) \cite{Davis97}, homotopy \cite{Osborne00}, least angle regression (LARS) \cite{Osborne00}, or soft-thresholding algorithms \cite{Daubechies04,Beck09}. 
Here we solve \eqref{l1} using a Generalized Lagrangian Multiplier Algorithm (GeLMA)~\cite{Moscoso12}.

In the next step, if $\bfX$ is known, the minimization problem for $\bfD$
\begin{equation}
\label{dict}
\begin{array}{rrclcl}
\displaystyle \min_{\bfD} & \multicolumn{3}{l}{\| \bfD \bfX - \bfY\|_F^2 }\, ,\\
%\textrm{s.t.} &\|\bd_k\|_2 =1,  k=1,\dots,K \\
\end{array}
\end{equation}
can be easily solved. The exact solution is $\bfD=\bfY\bfX^T(\bfX\bfX^T)^{-1}$,
provided $\bfX\bfX^T$ is invertible (actually stably invertible). % or that X be of full rank K=min{M,K}.
%all columns of $\bfX$ are linearly independent. 
%; just take the derivative of the objective function to obtain $( \bfD\bfX- \bfY) \bfX^T=0$. 
We can normalize the columns of $\bfD$  to one after they have been computed.

To summarize, in order to solve \eqref{dl_l1}, we alternate between problems \eqref{l1} and \eqref{dict}  to update $\bfX$ and $\bfD$ sequentially one after the other. %
% CHRYS: I commented this out 
Each iteration has two steps, starting from an initial guess $\bfX_0$  and $\bfD_0$. At the beginning of iteration $ l \geq 1$ we have $\bfD_{l-1}$ and $\bfX_{l-1}$, and we 
solve~\eqref{l1} with  $\bfD=\bfD_{l-1}$ to obtain $\bfX_{l}$ using GeLMA. Then, we solve \eqref{dict} with $\bfX=\bfX_{l}$  fixed to find %update $\bfD_{l-1}$ to 
\begin{equation}
\label{l2_inversion}
\bfD_{l}=\bfY\bfX_{l}^T(\bfX_{l}\bfX_{l}^T)^{-1}
\end{equation}
in the second step. 
This is very much like the Method of Optimal Directions (MOD) algorithm proposed in \cite{Engan00} for signal compression that uses Matching Pursuit for finding $\bfX_{l}$ instead of GeLMA in the first step.

%The approach consisting of two steps per iteration is common in other algorithms for dictionary learning. One of the most popular ones is the K-SVD algorithm \cite{Aharon06}, a generalization of the K-Means algorithm. %that solves \eqref{dl_l0} for $s=1$. 
%The K-SVD algorithm updates the dictionary using a spectral procedure on the residual and combines it with an update of the sparse representations to accelerate convergence. It is more efficient than MOD but its performance can compromise the final result as 
%the dictionary update given by MOD is the best possible for a fixed $\bfX$.
%We observed numerically that MOD performs better than K-SVD for the cases examined here. 
%Other methods that come with provable guarantees are \cite{Spielman12, Arora13, Agarwal17, Novikov23}. 

Our numerical experiments indicate that, given a suitable initialization, this dictionary learning algorithm can construct the matrix of Green's functions for wave propagation in random media. 
However, the columns of this matrix, the dictionary, are unordered. They cannot be used for imaging because we do not know the points where they focus in the image window. The next Section addresses this problem. 

\section{Grid reconstruction}
\label{sec:grid}
In this Section we describe an algorithm for finding the focal spots in the image window from the {\em estimated} Green's function vectors $\{\hat{\vect g}_i\}_{i=1}^{K}$. It is the range-free or connectivity based sensor localization algorithm \cite{Shang03}, analyzed in \cite{Oh10}. Our main contribution here is to determine connectivity  from cross-correlations of the Green's function vectors $\{\hat{\vect g}_i\}_{i=1}^{K}$, which now must retain some coherence.  However, the dictionary learning algorithm of the previous section requires incoherence, which means that the $\nu$ in (\ref{coh}) for the estimated $\{\hat{\vect g}_i\}_{i=1}^{K}$ is small. By using a subset of the components of the $\{\hat{\vect g}_i\}_{i=1}^{K}$, corresponding to a subarray as illustrated in Figure \ref{fig:schematic}, we increase their coherence. In this Section we use the same notation $\{\hat{\vect g}_i\}_{i=1}^{K}$ for their subsampled components.

This increased coherence is essential for the connectivity-based localization algorithm because it allows us to introduce a graph $G = (V,E)$ where the vertex set $V=\{1,2,\dots,K\}$ is associated with the estimated Green's function vectors 
$\{\hat{\vect g}_i\}_{i=1}^{K}$. A pair of vertices $(i,j)$ is then connected by an edge in $E$, and assigned the value one in the adjacency matrix of the graph, if the cross-correlation %\sout{$\vect g_i^* \vect g_j$}  
$\hat{\vect g}_i^* \hat{\vect g}_j$ is sufficiently close to one in absolute value. Otherwise, the pair $(i,j)$ is not connected, and zero is entered in the adjacency matrix. The size of the subarray of receivers is adjusted so that each vertex has up to $k=2r$ edges,
 where $r=2,3$ is the ambient dimension of the image window, assumed known. This is special for our imaging setup and could be generalized depending on the behavior of the cross-correlations.

The proxy distance between two Green's function vectors is now the geodesic graph distance between their corresponding vertices. That is, the proxy distance between $ \hat{\vect g}_i$ and $\hat{\vect g}_j$, denoted by $\hat{d}_{ij}$, is the number of edges in the shortest path connecting $i$ and $j$. 
We use this proxy distance as a replacement of the Euclidean distance between pairs of focal points in the image window associated with Green's function vectors in the MDS algorithm. The resulting configuration of focal points  $\hat{Z}=[\hat{\vx}_1, \hat{\vx}_2,\dots, \hat{\vx}_K]^T$ in the image window provides an estimate for the true configuration of focal points $Z=[\vx_1, \vx_2,\dots, \vx_K]^T$, up to rotation, translation and scaling.
%the MDS approach with cross-correlations 
%is summarized in Algorithm \ref{algo}. 
This is the MDS-MAP algorithm \cite{Shang03} with our correlation-based proxy distance. 
%For performance analysis of the MDS-MAP algorithm, we refer the reader to \cite{Oh10}. 
\begin{algorithm}
\begin{algorithmic}
\STATE {\bf INPUT:} $N\times K$ matrix $\hat{\vect G}$ with columns $\hat{\vect g}_i$, space dimensions $r=2,3$.
\STATE {\bf OUTPUT:} Matrix $\hat{Z}$ whose column vectors are the estimated coordinates of the grid points $\hat{\vx}_i$, $i=1,\ldots,K$.
\STATE
\STATE {\bf Compute}  $G = (V,E)$, with  $V=\{1,2,\dots,K\}$ and $E$ so that each node is connected to $2r$ neighbors; those corresponding to the 2r-largest values of  
$| \hat{\vect g}_i^* \hat{\vect g}_j |$.
\STATE {\bf Compute}  the proxy for distance matrix $\hat{D}$:
\vspace{-0.5cm}
\begin{quote}
 \IF{$(i,j) \in E$}
			\STATE $\hat{d}_{ij}=1$.
		\ELSE
			\STATE $\hat{d}_{ij} = $ shortest path along $G$
		\ENDIF
\end{quote}		
 \STATE {\bf Compute} $P = - {1 \over 2}L \hat{D} L^{T}$, where $L= \mathbf I_K - {\bf 1}_K  {\bf 1}_K^T/K$.
%and $\mathbf I_K$ is the $K\times K$ identity matrix.
 \STATE {\bf Diagonalize} $P$: $P=U\Sigma U^T$.
 \STATE {\bf Compute} $\hat{Z} = U_r\Sigma_r^{1/2}$.
%\STATE {\bf Compute} the proper roto-translation transformation and scaling to superimpose $X$ over the two anchors.
\end{algorithmic}
\caption{Reconstruction of focal points in image window}
\label{algo}
\end{algorithm}
%}

%In the algorithm, we set the distance between neighboring nodes to be equal to one and, then, we compute all pairwise distances for $(i,j) \notin E$  using the shortest path along the graph $G$. 
%This can be done with a greedy approach by computing the distances along all the possible paths that connect two nodes and selecting among them the path with the minimal distance. This is the geodesic distance on $G = (V,E)$, 
%{\em i.e.}, the distance between two vertices is the length \textcolor{red}{$\hat{D}$} of the path with the minimum number of edges. 

When the true Euclidean distance $D=(d_{ij})$ is used instead of $\hat{D}=(\hat{d}_{il})$ the classical metric MDS algorithm \cite{Borg} recovers the configuration of focal points $Z=[\vx_1, \vx_2,\dots, \vx_K]^T$ up to rotation and translation. In this case
the input is a $K\times K$ squared distance matrix  $D$ with entries $d_{ij}$,
 $d_{ij}=(\vx_i - \vx_j)^T(\vx_i - \vx_j)$, 
and the output the  $K\times r$ configuration matrix of focal points 
%\sout{$Z=[\vx_1, \vx_2,\dots, \vx_K]^T$} 
$Z$. We have that \cite{Borg}
\begin{eqnarray}
\label{matrixP}
- {1\over 2} L D L= L Z Z^T L \, ,
 \end{eqnarray} 
 where  $L= \mathbf I_K - {\bf 1}_K {\bf 1}_K^T/K$ is a centering matrix, with $\mathbf I_K$ the $K\times K$ identity matrix, and  ${\bf 1}_K$ the column vector of all ones. This means that the matrices $Z Z^T$ and $-D/2$ are equal when the center of mass of the configuration is moved to zero. For the Euclidean distance matrix $D$ the focal point reconstruction algorithm (\ref{algo}) determines the Euclidean coordinates of the focal points. 
% a rigid transformation (translations, rotations, and reflections). 
%The matrix $P$ is symmetric, and, thus, its singular value decomposition  $P=U\Sigma U^T$ determines the coordinates 

%The relation of neighborhood is is symmetric and for r = 2 each node has four neighbors while 317 for r = 3 each node has six neighbors.
 %to the $4$ (resp. $6$) 

%The relation of neighborhood is symmetric and for $r=2$ each node has four neighbors while for $r=3$ each node has six neighbors. 

\begin{figure}[htbp]
\centerline{
    \includegraphics[width=0.41\linewidth]{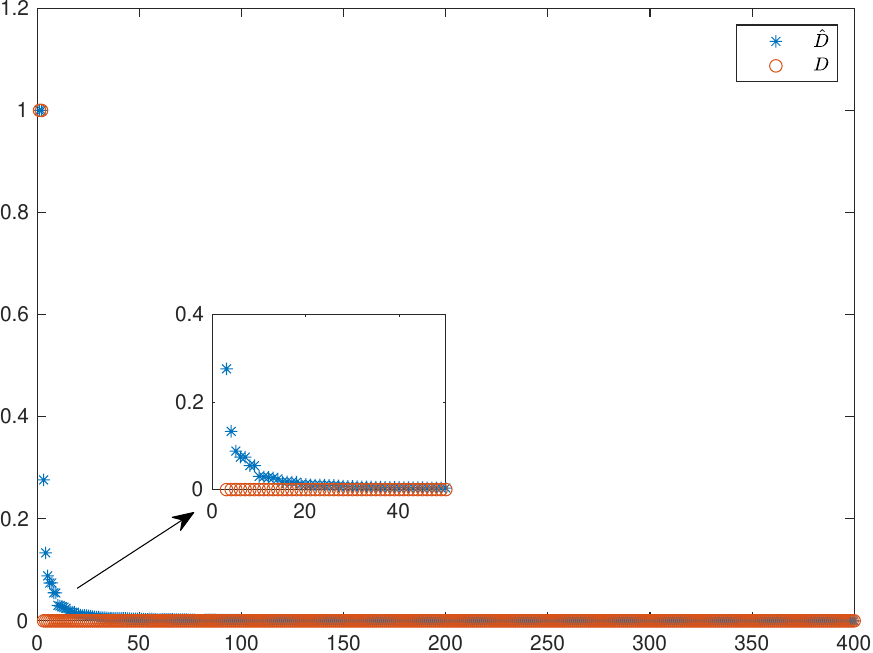}
  }
\caption{%Adding randomness helps the MDS-MAP algorithm. Grid reconstructed using the MDS-MAP algorithm with only four nearest neighbors (left) and with all pairwise distances (center). 
The singular values of the doubly centered distance matrix $P$ normalized by the maximal one are plotted with red circles when $D$ is used (rank is exactly 2) and with blue stars when $\hat D$ is used. There are exactly 2 top singular values in the second case as well, plotted with blue stars, and the lower eigenvalues drop to zero fast but are not immediately zero as with the red circles.}
%\textcolor{red}{Can we use an insert on the right panel to show the knee of the curve near the origin? Should this figure be placed earlier in the paper?}
    \label{fig-grid-rand}
    \end{figure}

%It is well understood that the MDS-MAP algorithm provides an approximation of the location of the nodes, that is the matrix $Z$, up to centering. 
The rank of the matrix $P=- {1\over 2} L D L$ equals the ambient dimension $r$ of the image window when the Euclidian distance matrix $D$ is used in the MDS algorithm. When we use the geodesic distance $\hat D$ on the graph then the rank of $P$ is not equal to $r$ any more. However, the first $r$ singular vectors of $P$ are close to the true coordinates $Z$ (up to centering and rotation) \cite{Oh10}. This is illustrated in 
%\sout{the right panel of } 
Figure \ref{fig-grid-rand}.
%\textcolor{red}{(I would move this figure in the text file so it becomes Fig. 3 instead of Fig. 7)}.
%of the singular values of $P$ when using $D$ and $\hat D$ is shown in Figure \ref{fig-grid-rand} (right).
% where the singular values of $P$ are plotted normalized by their maximal one. 
The absolute location of the focal points in the image window can be determined
using the true location of a few of them, the anchors. These anchors allow us to find the proper rigid transformation and scaling to superimpose the given configuration over them. The anchors can be known a priori or their location can be estimated using coherent interferometric imaging \cite{Borcea11}.  The number of anchors needed is small, typically $r+1$. % 

\section{Numerical experiments}
\label{sec:experiments}

To simulate wave propagation in random media we use the random travel time model (\cite{Borcea11,Moscoso17} and references therein) which provides an analytical approximation for the Green's function in \eqref{eq:wave equation} in the high-frequency regime in random media with weak fluctuations and large correlation lengths $\ell=100\lambda$ compared to the central wavelength $\lambda$, given by
\begin{equation}\label{eq:random green func}
\wG(\vx,\vy)=\wG_0(\vx,\vy)
\exp{\left[ i \sigma\kappa|\vx-\vy|\int_0^1\mu(
\frac{\vx}{l}+\frac{s}{l}(\vy-\vx))\, ds\right]}\, .
\end{equation}
Comparing Eqs. (\ref{eq:random green func}) and (\ref{greenfunc}) for a homogeneous medium we see that, in this regime, only the phases are perturbed by the random medium while the magnitudes remain unchanged.

\begin{figure}[htbp]
\centerline{
  \includegraphics[width=7cm]{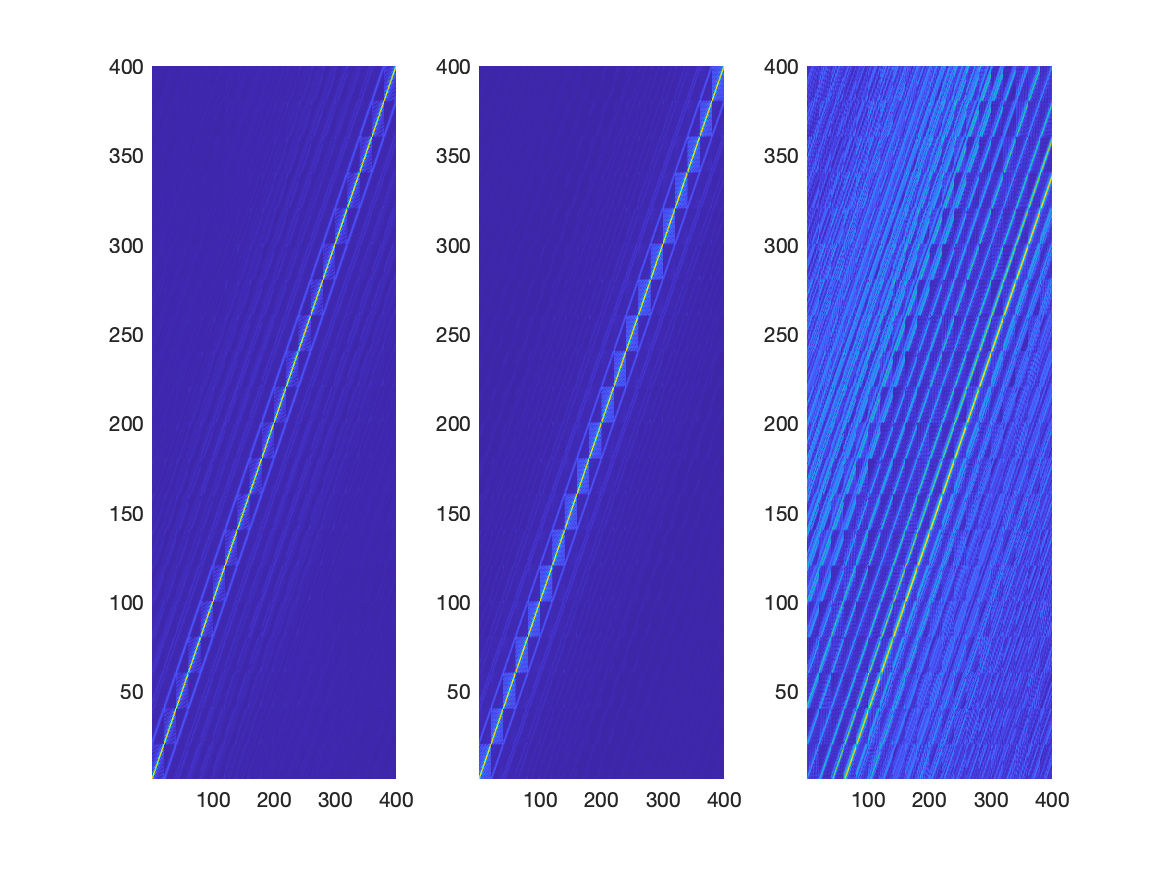}}
  \caption{Cross-correlations of the sensing matrix. Left: cross-correlations of the sensing matrix in the random medium. Center: cross-correlations of the sensing matrix in the homogeneous medium. Right cross-correlations between the sensing matrix in the random and homogeneous media. Strength of the fluctuations of the random medium $\tilde{\sigma}=0.8$. }
    \label{fig_correl}
\end{figure}

In our numerical experiments the distance between the array and the image window $L=100 \ell$ is large,
so the small distortions produced by each inhomogeneity build up over the propagation distance and are significant at the receivers. The strength of the fluctuations $\sigma$ is scaled by the dimensionless parameter $\lambda/\sqrt{lL}$, for which the standard deviation of the random phase fluctuations in the Green's function is $\mathcal{O}(1)$.   The strength of the fluctuations $\tilde{\sigma} = \sigma / (\lambda/\sqrt{lL})$ in the simulations is   $\tilde{\sigma} =0.6$ or $\tilde{\sigma} =0.8$. 
%Here $\ell$ denotes the correlation length $\ell=100\lambda$ with $\lambda$ defined with respect to the frequency $f$, $\lambda = c/f$.

\begin{figure}[htbp]
      \includegraphics[width=4.2cm]{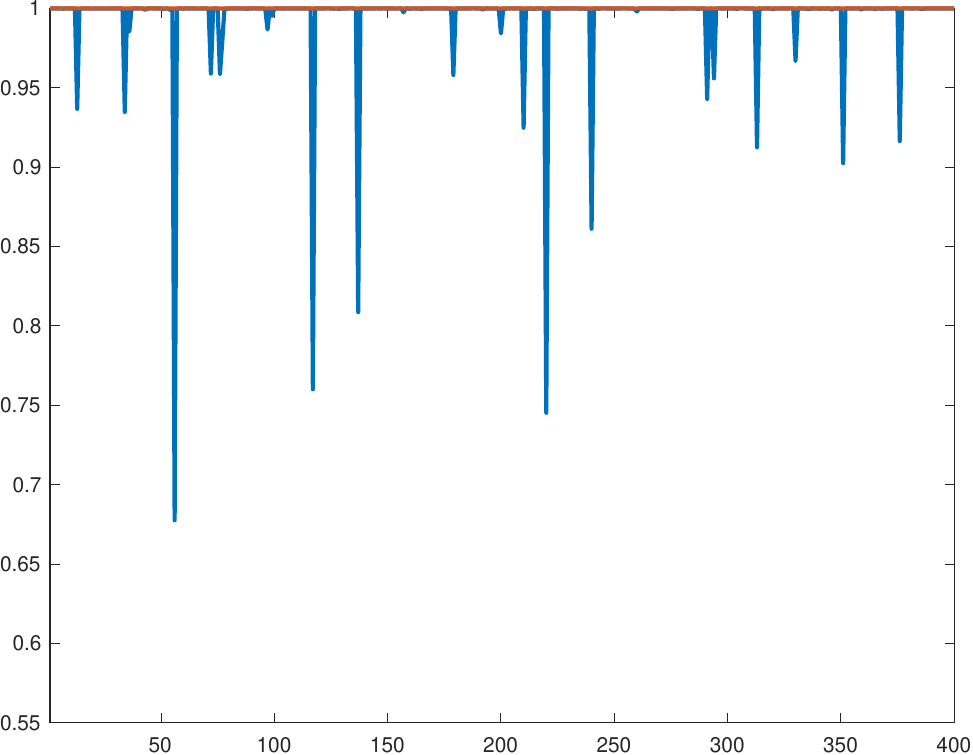}
      \includegraphics[width=4.2cm]{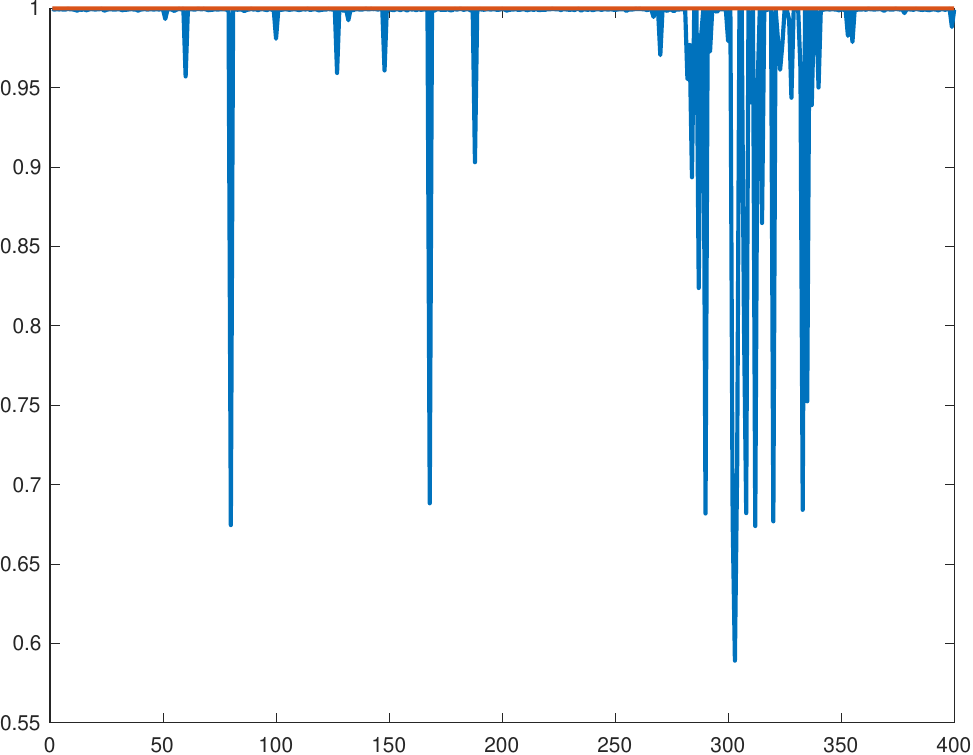}
 %}
  \caption{Maximum correlation between each estimated column and the true ones in the sensing matrix, as in (\ref{compDL}). In red the results when the columns of the matrix are more incoherent ($a=48 \ell $). In blue the results when the columns of the matrix are more coherent ($a=24 \ell $). Sparsity $s=4$ on the left and $s=8$ on the right.  %Recovering the columns of the sensing matrix becomes more challenging as the sparsity increases. 
  }
    \label{fig-angle}
\end{figure}

We consider the following setup for our numerical simulations.  In the first step of dictionary learning for the sensing matrix $ \vect{\Gc}$ we use the multi-frequency data recorded with a large array aperture $a=48 \ell $ with $N_r=145$ equally spaced receivers (see Fig. \ref{fig:schematic}). In the second step for the grid reconstruction, we use the data corresponding to half the array aperture, so $a=24 \ell $. 
%discretized with $73$ equispaced receivers (see Fig. \ref{fig:schematic}). 
The bandwidth $[0.5 f, f]$, with $f={c_0}/\lambda$, is discretized with $N_f=10$ equally spaced frequencies, and is the same for both
steps of the algorithm. We  organize the multiple frequency data column-wise, so
$$\bfY= [\bfY(f_1)^{\intercal}, \bfY(f_2)^{\intercal},\dots,\bfY(f_{N_f})^{\intercal}]^{\intercal}$$
and the multi-frequency sensing matrix is now
$$\vect\wg_i= [\vect\wg(\vx_i,f_1)^{\intercal}, \vect\wg(\vx_i,f_2)^{\intercal},\dots,\vect\wg(\vx_i,f_{N_f})^{\intercal}]^{\intercal},$$ 
$i=1,\dots,K$. Thus, the sensing matrix $\vect{\Gc}=[\vect\wg_1\,\cdots\,\vect\wg_K]$ has dimensions $N\times K$ with
The sampling of the $20\times 20$ points in the image window is based on the homogeneous medium array resolution, $O(\lambda L/a)$ in cross-range and 
%\sout{$O(c/B)$}  
$O(c_0/B)$ in range \cite{Born70,Borcea02}.

In Figure \ref{fig_correl}, we assume that the sensing matrices corresponding to a random medium $\vect{\Gc}$ and 
to the homogeneous medium $\vect{\Gc}_0$ are known, and we show the cross-correlation matrices of $\vect{\Gc}^*\vect{\Gc}$ (left), $\vect{\Gc}_0^*\vect{\Gc}_0$ (center), and $\vect{\Gc}^*\vect{\Gc}_0$ (right). 
%The subscript $r$ denotes random medium, while the subscript $0$ is used for the homogeneous one. 
For the homogeneous medium the Green's function used is given by (\ref{greenfunc}). 
Each row $i$ in these images corresponds to a time reversal experiment where a source located at $\vx_i$ emits a pulse, and the recorded signals are time reversed and emitted back into the medium. 
When the waves are re-emitted into the same medium in which the measurements were obtained, as in the left and center images of Figure \ref{fig_correl},  they retrace the original scattering process and arrive back approximately at 
the  point at  which they were emitted, that is, the focal point. However, when the back-propagation is done in a different medium, as in the right image of this figure, there is no re-focusing. In Figure \ref{fig_correl},  
the large values (lighter blue color) correspond to re-focusing points. This figure shows that (a) time reversal of waves into random and homogeneous media are similar, and 
(b) that we cannot use the homogeneous medium to recover this structure if there is scattering. 

\begin{figure}[htbp]
\centerline{
  \includegraphics[width=0.25\linewidth]{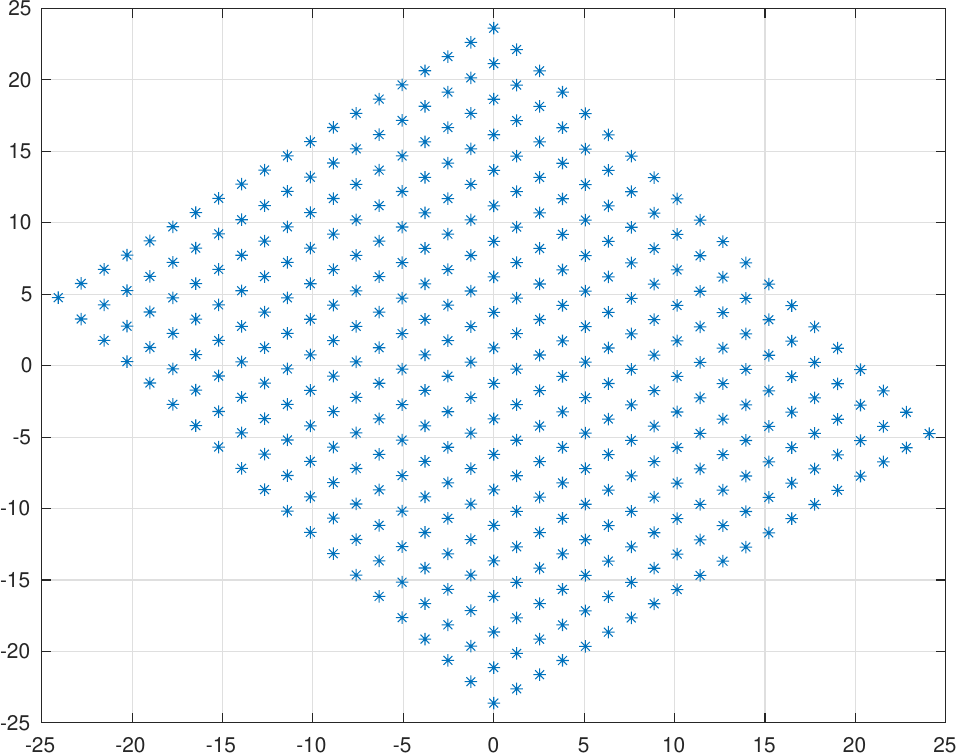}
  \includegraphics[width=0.25\linewidth]{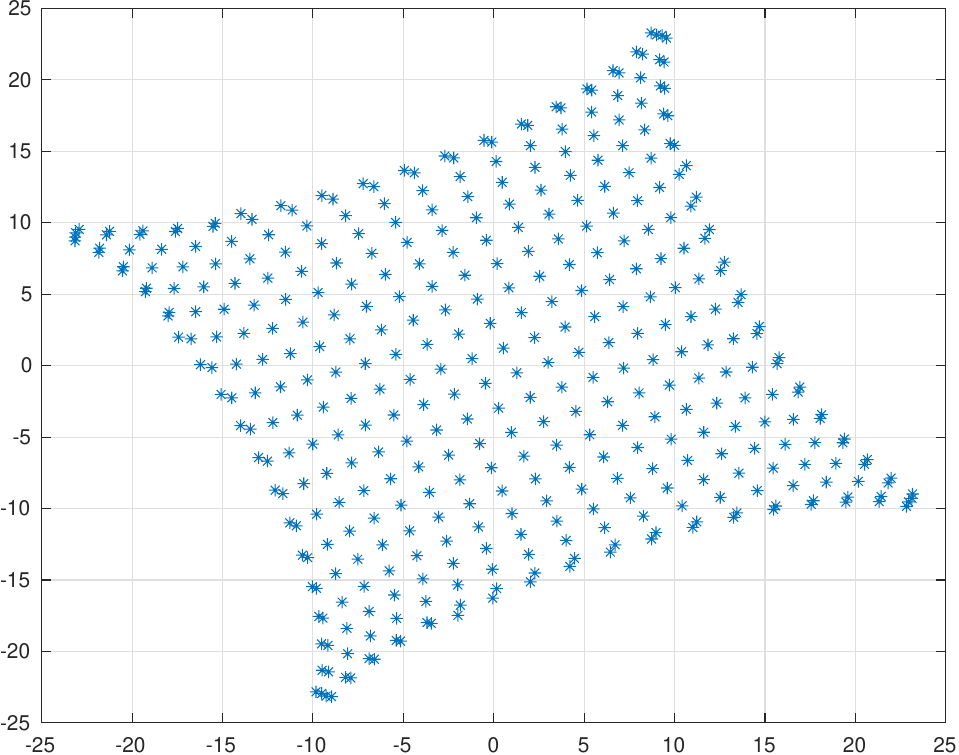}
    \includegraphics[width=0.25\linewidth]{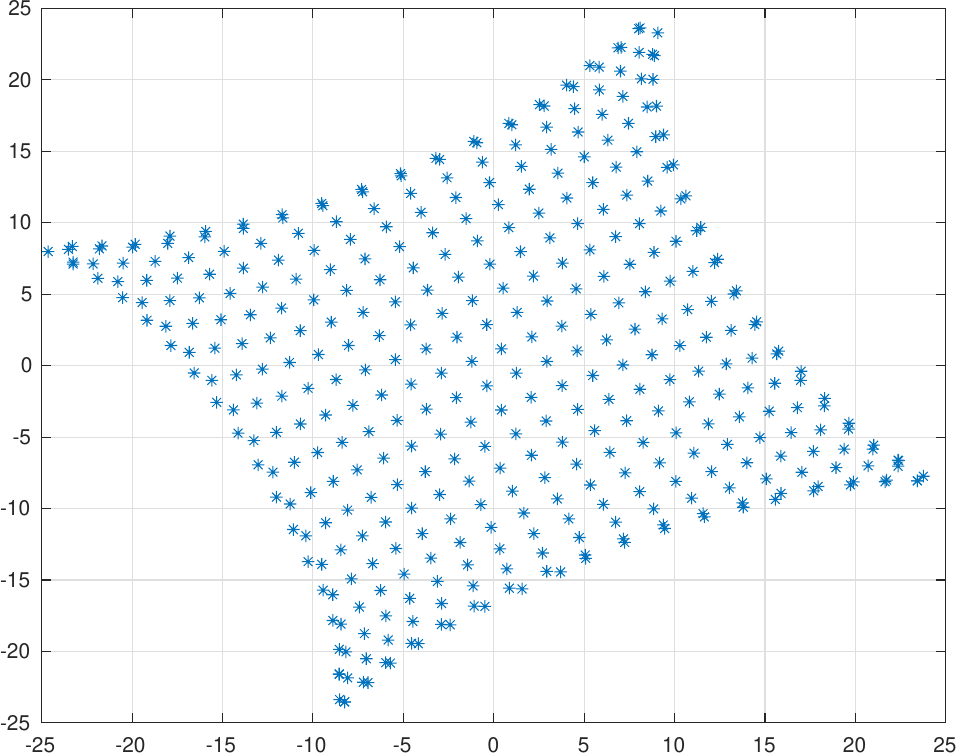} 
      \includegraphics[width=0.25\linewidth]{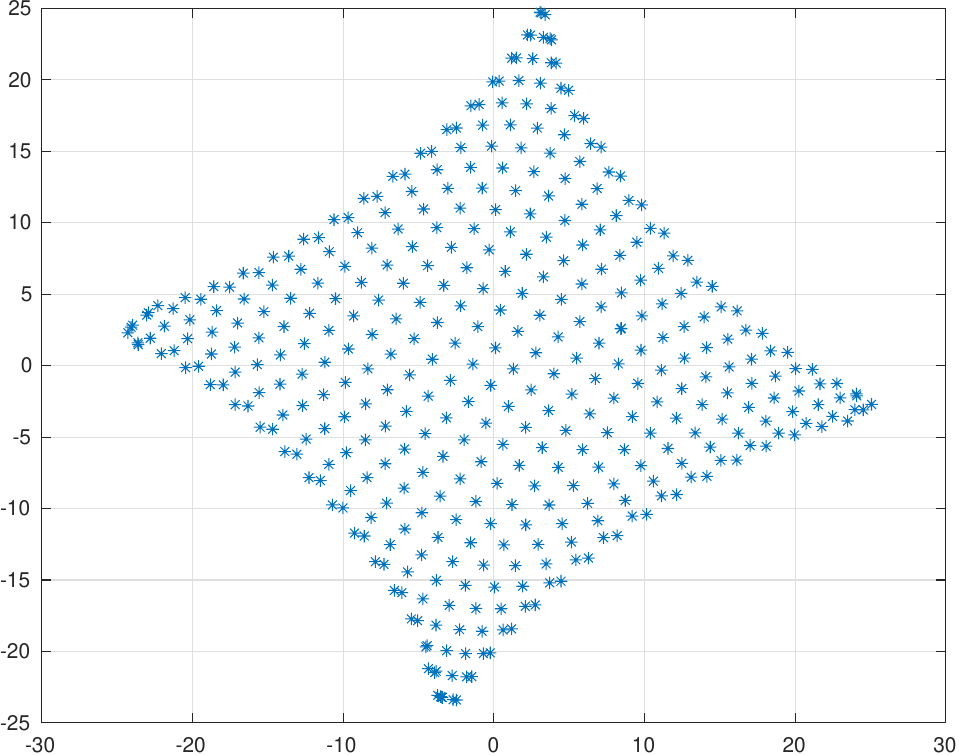} 
  }
  \caption{From left to right: Grid reconstruction from true Euclidean distances using the MDS algorithm when all the pairwise distances are assumed known;  from true Euclidean distances when only distances corresponding to the four nearest neighbors are assumed known; using the MDS-MAP algorithm with geodesic graph distances for $\tilde{\sigma}=0.6$; and using the MDS-MAP algorithm with geodesic graph distances for $\tilde{\sigma}=0.8$. Sparsity $s=8$ in all cases.}
    \label{fig-grid}
\end{figure}

In our numerical experiments, we assume that we have a diverse set of data $\bfY=[\by_1,\by_2, \ldots,\by_M]$, with 
$\by_i = \vect{\Gc} \, \bx_i$. Both the sensing matrix $\vect{\Gc} \in C^{N \times K}$ and the sparse vectors $\bx_i \in \mC^{K}$ are unknown.  We assume  data corresponding to a large number of experiments, so $M \gg K$.
Given this set of data, we want to recover the columns of the sensing matrix $\vect{\Gc} =[\vect\wg_1\,\cdots\,\vect\wg_K]$, whose rank is approximately 
$200<K=400$ and whose coherence is $\nu= 0.7$. The sensing matrix is rank-deficient because the resolution of the image window is high, with pixel sizes $\lambda L/a$ in cross-range and $c_0/B$ in range.

The results of the first step of the proposed strategy are depicted in Figure \ref{fig-angle} for large (red lines) and small (blue lines) arrays. We solve the problem (\ref{dl_l1}) as described in Section \ref{sec:dl}. 
To measure the success of this first step  we form, for every (normalized) recovered column $\hat{\vect\wg}_i$, the cross-correlations with all the  columns of  the true sensing matrix $\vect{\Gc}$, and  represent in Figure \ref{fig-angle} the maximum value
\begin{equation}
\label{compDL}
C_{max}(i)= \max_j | \hat{\vect\wg}_i^T \vect\wg_j |
\end{equation}
for a sparsity level $s=4$ (left) and $s=8$ (right).
We observe
values very close to $1$ in both cases when the columns of the sensing matrices are for large array apertures (red lines). This means that the true Green's function vectors are recovered when  large aperture arrays are used because they are incoherent. However, when smaller arrays are used (blue lines) Green's function vectors are coherent and some of them are not recovered. It is very important to recover accurately all, or almost all, Green's function vectors because, otherwise, we cannot establish their connectivity properly and, therefore, we cannot reconstruct the grid in the image window in the second step.

\begin{figure}[htbp]
\centerline{
    \includegraphics[width=0.3\linewidth]{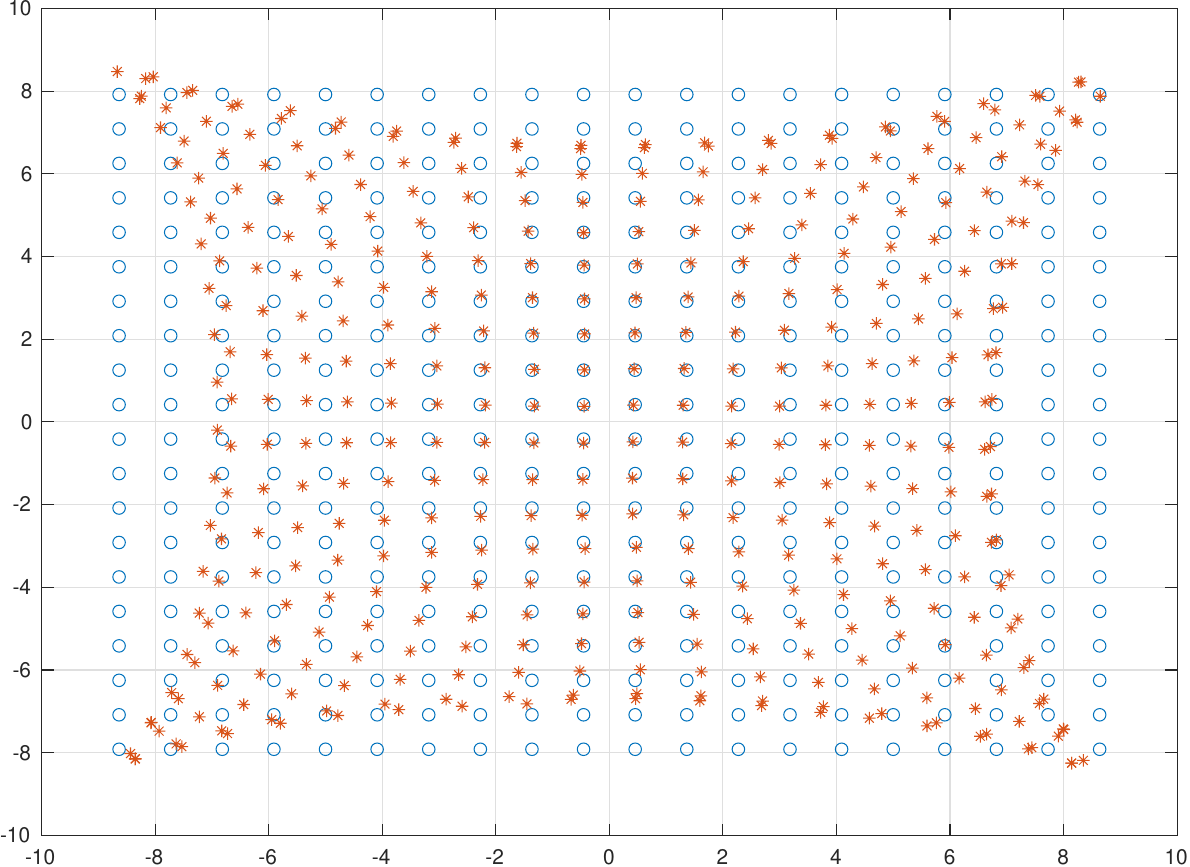} 
     \includegraphics[width=0.28\linewidth]{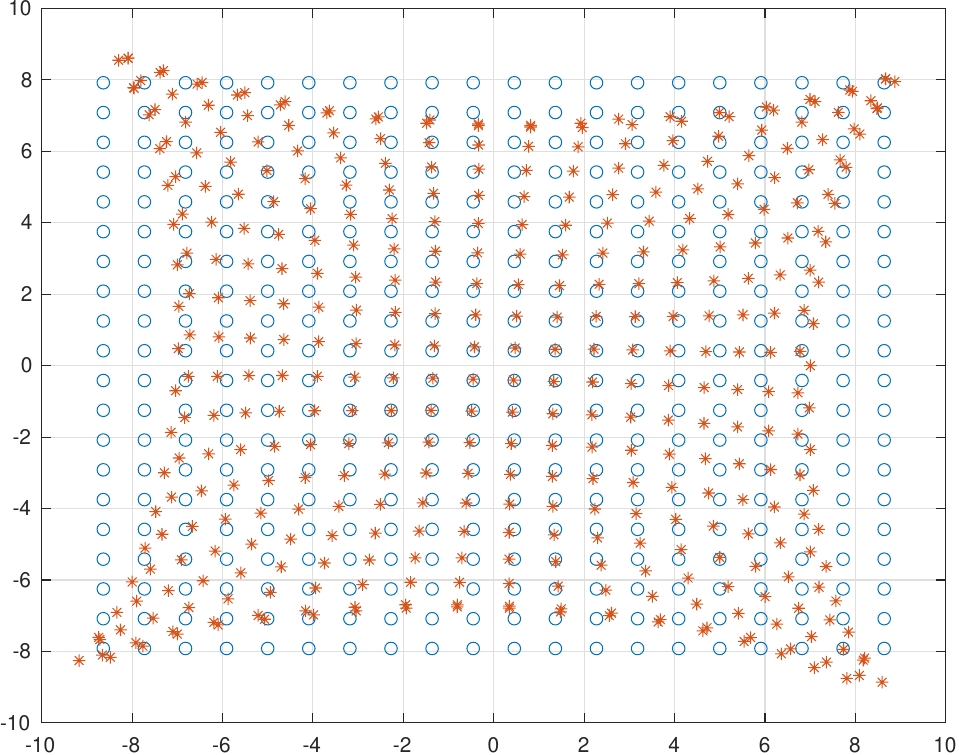} 
  }
  \caption{Using two points as anchors, {\em i.e.} assuming the location of those two points is known, we can estimate the scaling and the rotation needed to recover the absolute grid positions. We compare the recovered locations (red stars) with the true ones (blue circles) where $\tilde{\sigma}=0.6$ (left) and $\tilde{\sigma}=0.8$ (right).
Sparsity $s=8$ in both cases.}
    \label{fig-grid2}
\end{figure}

Figure \ref{fig-grid} shows the results for the grid reconstructions accomplished with algorithm (\ref{algo}) described in Section \ref{sec:grid} using $k=4$ neighbors.
This algorithm provides the correspondence between the Green's function vectors found in the first step and their focal points in the image window.
From left to right we show the results when (left) all the pairwise Euclidean distances between the focal points are known in a homogeneous medium, (second from the left) when only Euclidean distances between the four nearest neighbors are known in a homogeneous medium, (second from the right) using only connectivity information in a random medium with $\tilde{\sigma}=0.6$, and  (right) using only connectivity information in a random medium with  $\tilde{\sigma}=0.8$. In all the cases, the sparsity level is $s=8$. The algorithm (\ref{algo}) provides grid positions up to a rigid transformation and scaling. We post-process the results shown in the second from the right and right images in Figure \ref{fig-grid} to transform them to absolute positions using two anchors. See Figure \ref{fig-grid2}.

We observe in Figure \ref{fig-grid2} that the grids are quite well reconstructed near the center but bent towards the edges. This occurs because our geodesic graph distance is the
scaled $l_1$ distance on the grid, and an embedding of such distances into Euclidean spaces leads to such distortions. 
%As illustrated in Figure \ref{fig-grid}-left similar results are obtained even when the true distance for the four nearest neighbors for each point is used. 
Naturally, there is no grid deformation shown in the left image of Figure \ref{fig-grid} since Euclidean distances are used. %when all pairwise distances are used in the MSD algorithm. 

\begin{figure}[htbp]
  \centering
  \includegraphics[width=0.32\linewidth]{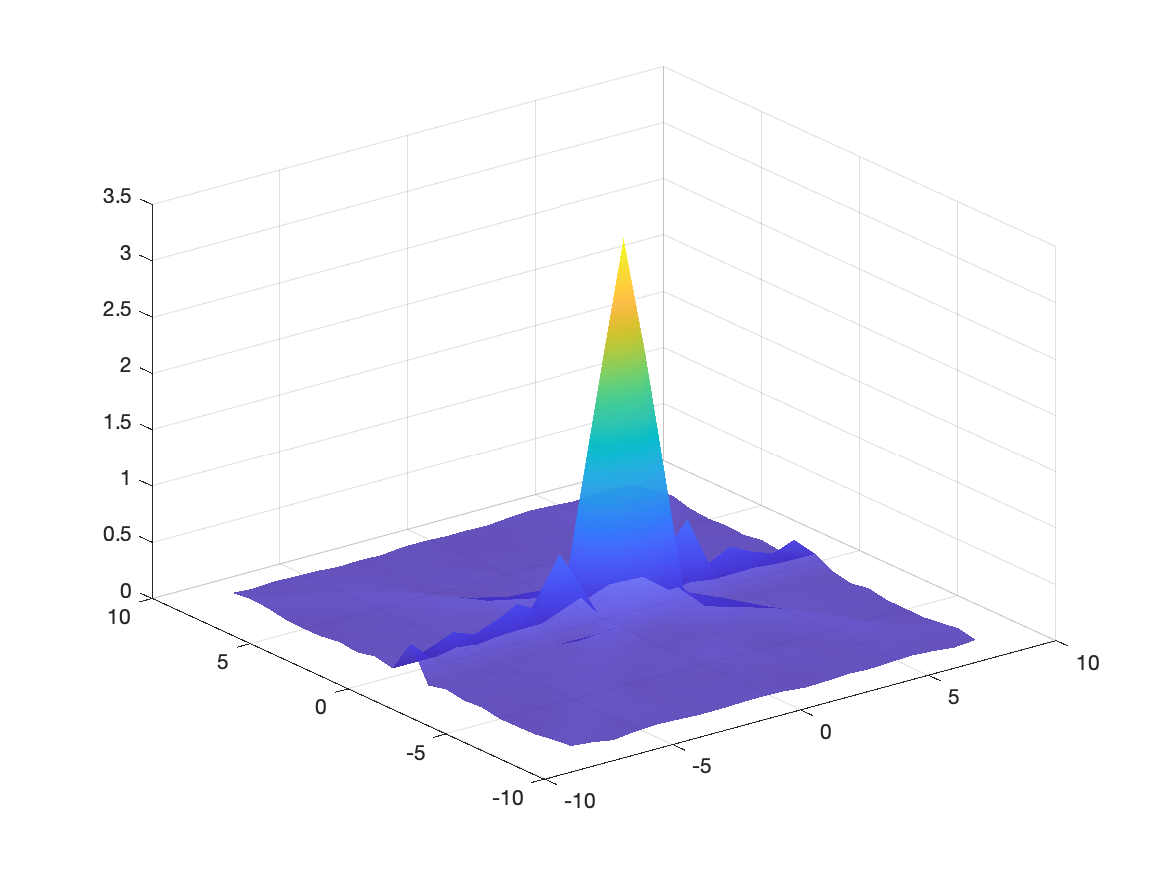}
  \includegraphics[width=0.32\linewidth]{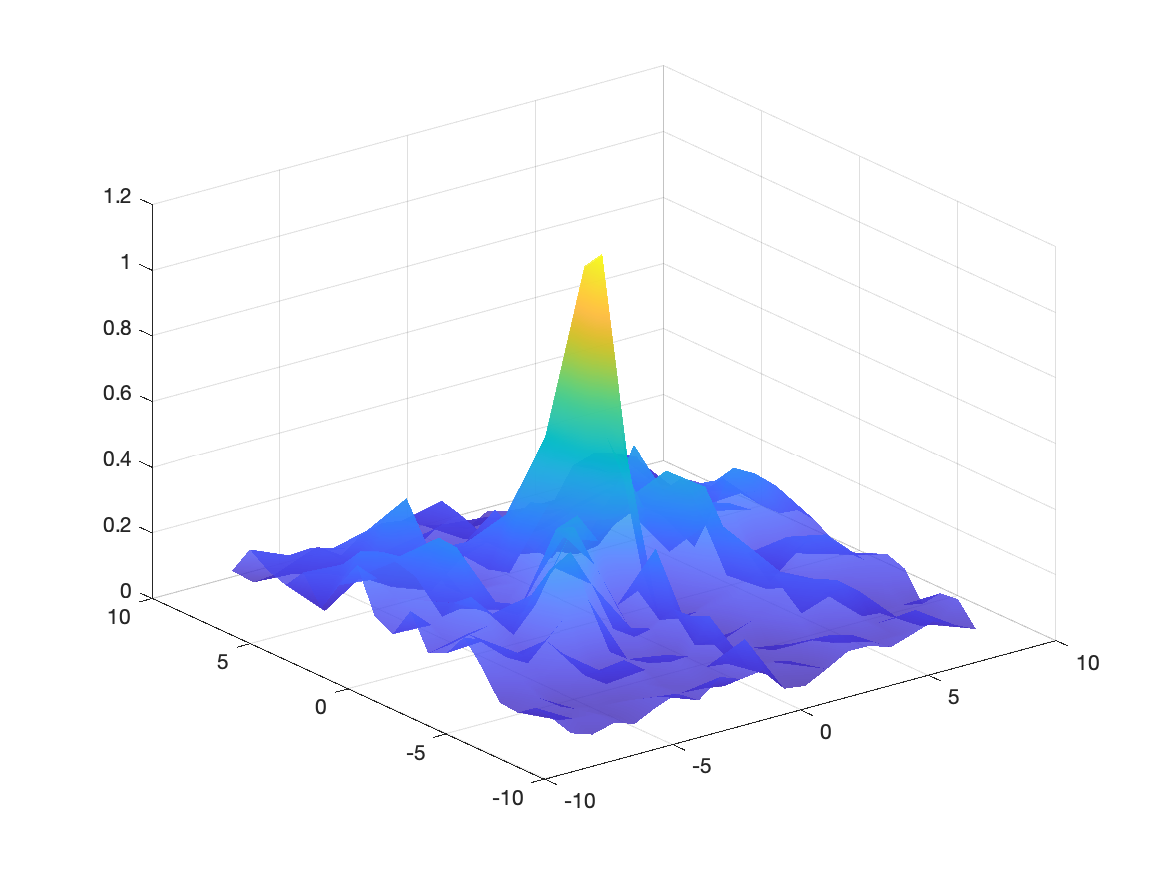}
  \includegraphics[width=0.32\linewidth]{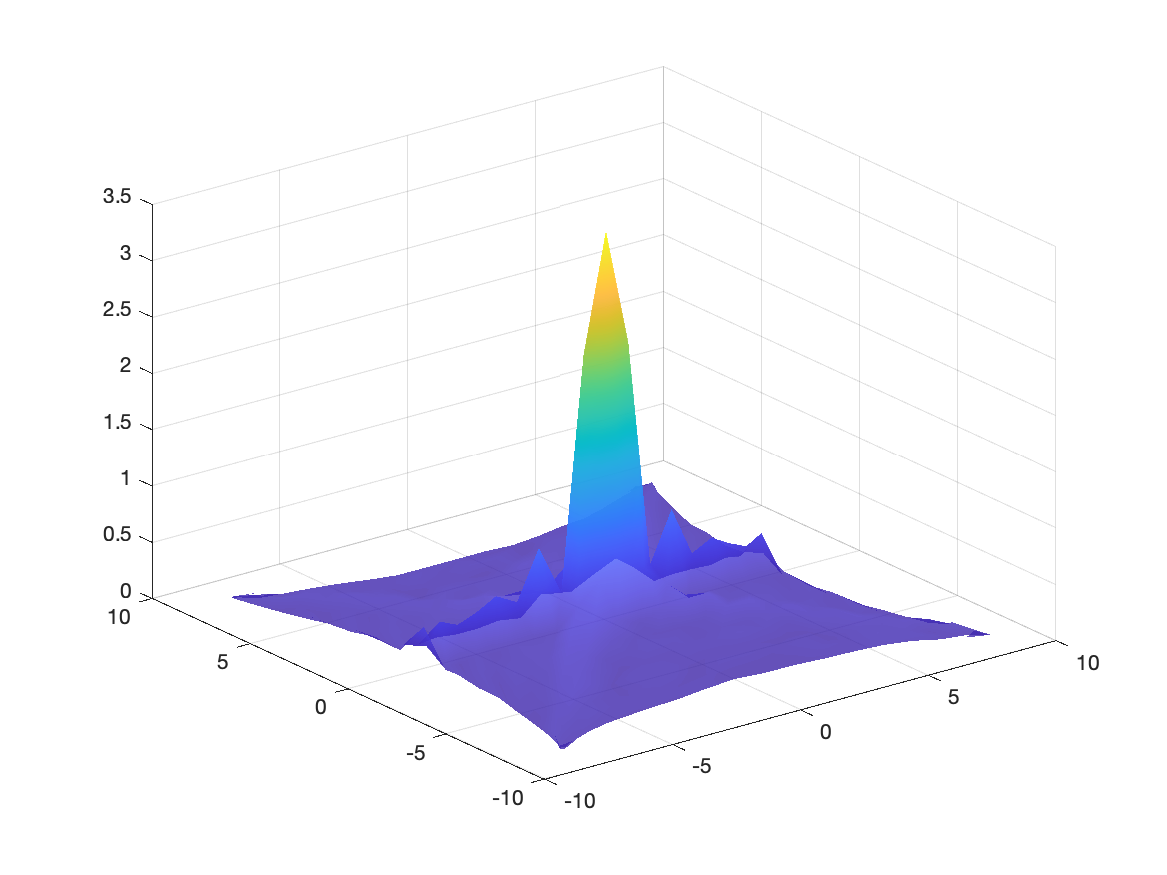}\\
   \includegraphics[width=0.32\linewidth]{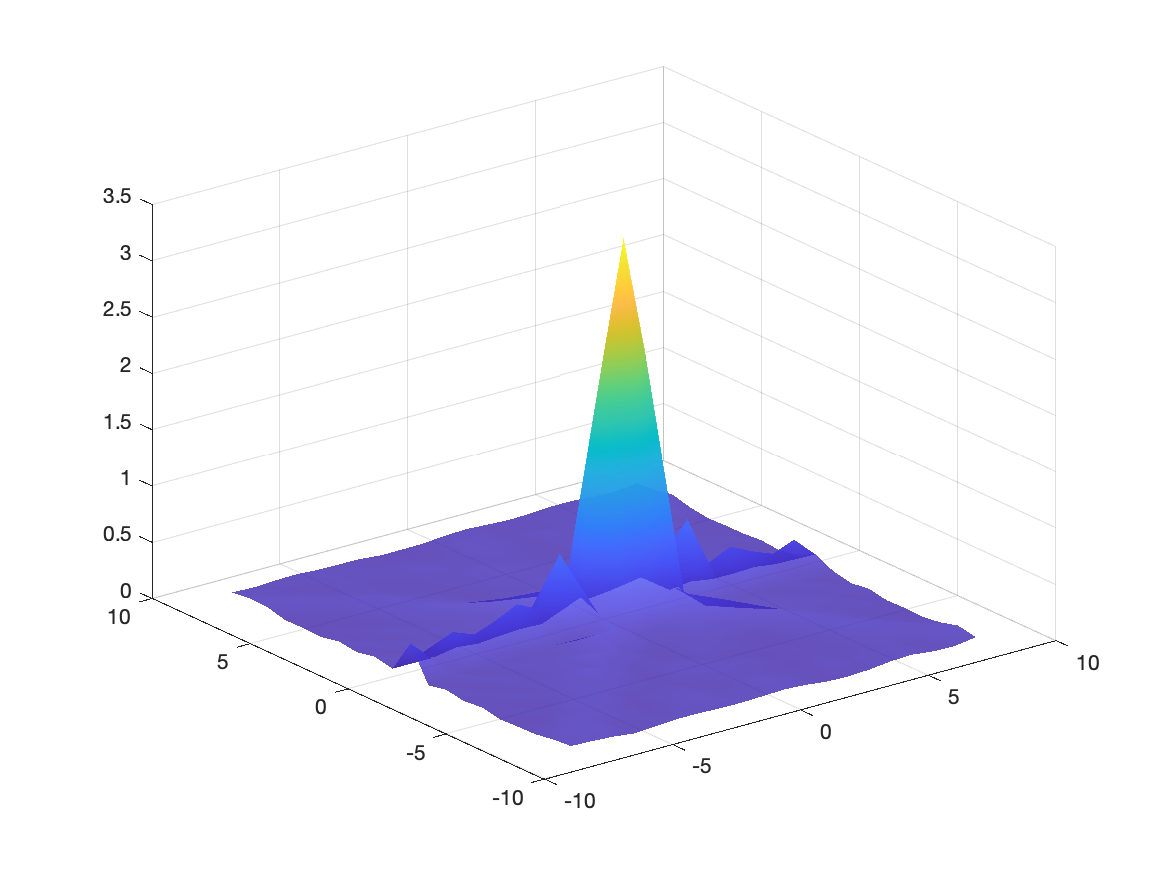}
  \includegraphics[width=0.32\linewidth]{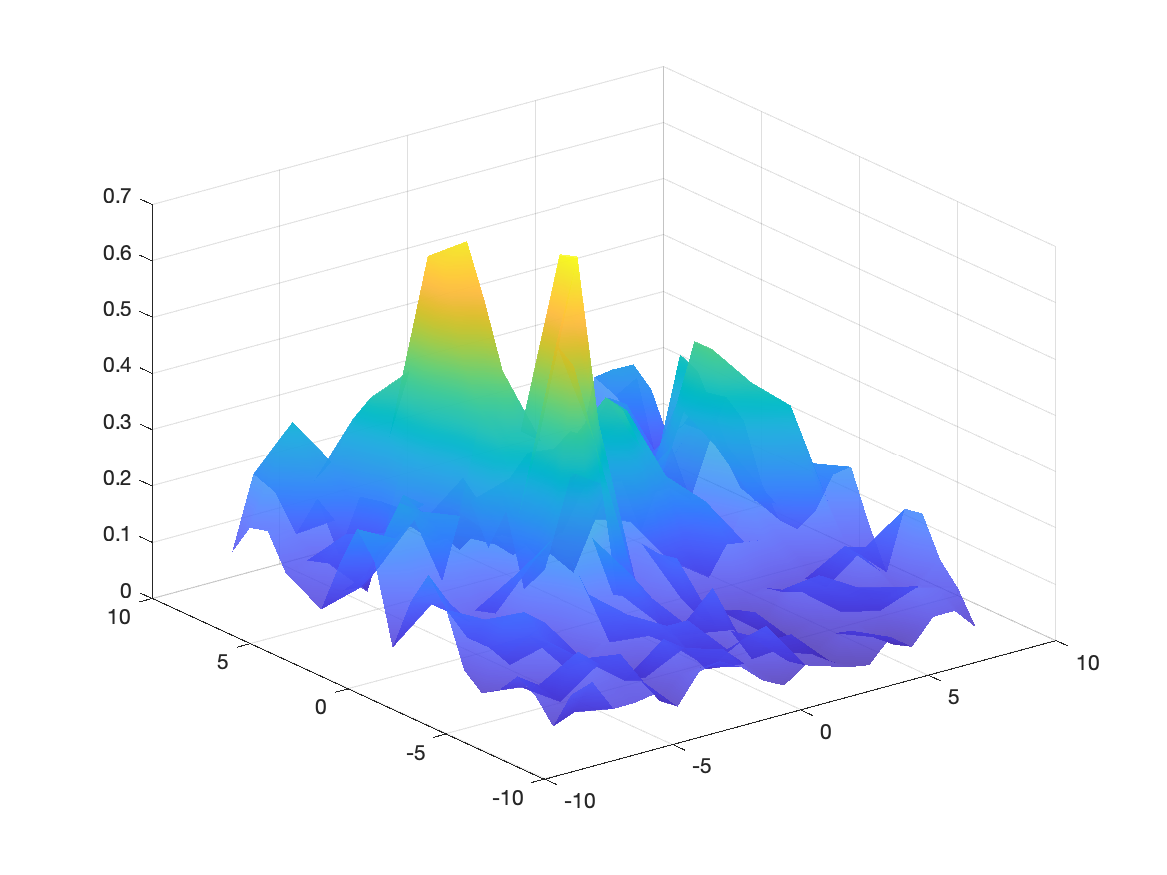}
  \includegraphics[width=0.32\linewidth]{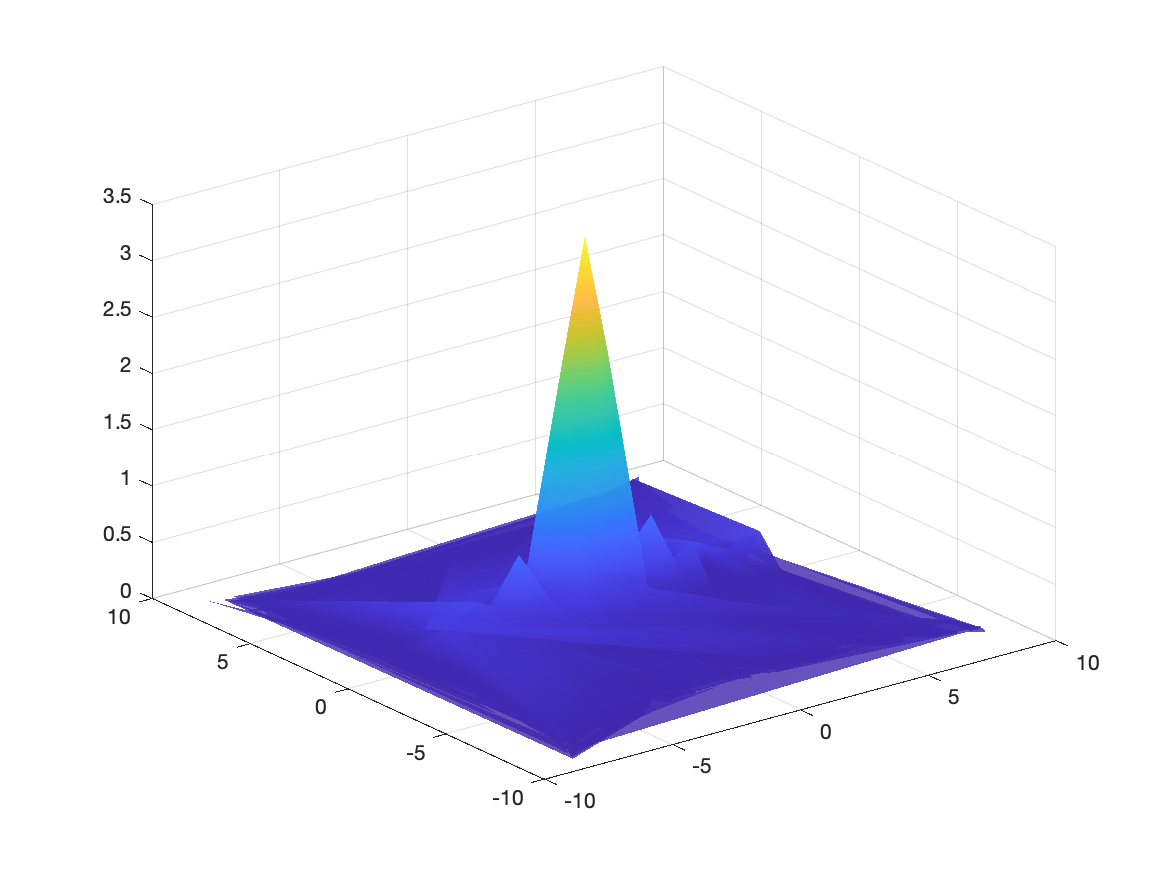}
  \caption{From left to right, image formed with (\ref{eq:imag}) using the true random Green's functions, the homogeneous Green's functions and the recovered ones with the proposed method. Here the sparsity is $s=8$. The strength of the fluctuations is $\tilde{\sigma}=0.6$ for the top row and 
 $\tilde{\sigma}=0.8$ for the bottom row. }
  \label{fig_new}
\end{figure}

After the two steps of the proposed strategy, we recover the ordered sensing matrix $\hat{\vect{\Gc}}$. We can, therefore, image any signal that comes to the array from the image window. In Figure \ref{fig_new}, we back-propagate a signal $\vect\wg(\vx_j)$ from a source located at  $\vx_j$ using the recovered Green's function vectors, so the image formed at points $\vx_i,\ i=1,\ldots,K$ is 
\begin{equation}
\label{eq:imag} 
{\mathcal I}(\vx_i; \vx_j)=  \left|  \hat{ \vect\wg(\vx_i)}^* \vect\wg(\vx_j)  \right|.
%\left| \vect{\hat{\Gc}}^* \vect\wg(\vx_j) \right| 
\end{equation} 
%which takes the value $ \left|  \hat{ \vect\wg(\vx_i)}^* \vect\wg(\vx_j)  \right|$ . 
As before, the hat in \eqref{eq:imag}, denotes the recovered Green's function vectors of the sensing matrix using the two step method introduced here.
As illustrated in Figure \ref{fig_new}, the image produced (right) is similar to the one obtained using the true Green's functions (left) and significantly better than the one obtained using the homogeneous Green's function (center). 
This figure shows the need for recovering accurate estimates of the Green's function vectors for imaging in random media since the ones corresponding to a reference homogeneous medium provide very noisy, useless images with \eqref{eq:imag}.
The top and bottom rows are images obtained in different realizations of the random media with $\tilde{\sigma} =0.6$ and $\tilde{\sigma} =0.8$, respectively. 
%\textcolor{red}{This must be edited to account for "improved" resolution in RM because TR in RM focuses better. So we need to compare to a vanilla homogeneous medium}

%Engan, K.; Aase, S.O.; Hakon Husoy, J. (1999-01-01). Method of optimal directions for frame design. 1999 IEEE International Conference on Acoustics, Speech, and Signal Processing, 1999. Proceedings. Vol. 5. pp. 2443–2446 vol.5. doi:10.1109/ICASSP.1999.760624. ISBN 978-0-7803-5041-0. S2CID 33097614.

{\bf Acknowledgements} Miguel Moscoso's work was  supported by  the Spanish AEI grant PID2020-115088RB-I00.
Alexei Novikov's work was partially supported by  AFOSR FA9550-23-1-0352 and FA9550-23-1-0523. 
The work of George Papanicolaou was partially supported by AFOSR FA9550-23-1-0352.
The work of  Chrysoula Tsogka was partially supported by AFOSR FA9550-23-1-0352 and FA9550-21-1-0196.


\begin{thebibliography}{90}

\bibitem{Agarwal16} A. Agarwal, A. Anandkumar, P. Jain, and P. Netrapalli, {\em Learning sparsely used overcomplete dictionaries via alternating minimization,} SIAM Journal on Optimization {\bf 26}, 2775--2799 (2016). 
%
\bibitem{Agarwal17} A. Agarwal, A. Anandkumar, and P. Netrapalli, {\em A clustering approach to learning sparsely used overcomplete dictionaries,} IEEE Transactions on Information Theory {\bf 63}, 575--592 (2017). 
%
\bibitem{Aharon06} M. E. M. Aharon, M. Elad and A. Bruckstein, {\em K-SVD: An algorithm for designing overcomplete dictionaries for sparse representation,}  IEEE Transactions on Signal Processing {\bf 54}, 4311--4322 (2006). % doi: 10.1109/TSP.2006.881199.
%
%\bibitem{Aharon06b}
%M. E. M. Aharon and A. Bruckstein, {\em On the uniqueness of overcomplete dictionaries, and a practical way to retrieve them,} Linear Algebra and its Applications {\bf 41}, 48--67 (2006).
%%doi: 10.1109/TIT.2016.2614684.
%
%\bibitem{Arora13} S. Arora, R. Ge, and A. Moitra, {\em New algorithms for learning incoherent and overcomplete dictionaries,} Journal of Machine Learning Research {\bf 35} (2013).
%
\bibitem{Aspnes07} J. Aspnes, T. Eren, D.K Goldenberg, W. Whiteley, Y.R. Yang, and B.D. O. Anderson,  A Theory of Network Localization, Mobile Computing, IEEE Transactions {\bf 5}, 1663--1678 (2007). %10.1109/TMC.2006.174. 
%
%\bibitem{Babcock53} H. W. Babcock, {\em The possibility of compensating astronomical seeing,} Publ. Astron. Soc. Pac. {\bf 31}, 229--236 (1953).
%
%%\bibitem{Badon20} A. Badon, V. Barolle, K. Irsch, A. C. Boccara1, M. Fink, A. Aubry, {\em Distortion matrix concept for deep optical imaging in scattering media,} Sci. Adv. {\bf 6}, 7170 (2020).
%
%
%\bibitem{Bardos08} C. Bardos, J. Garnier, and G. Papanicolaou, {\em Identification of Green's functions singularities by cross correlation of noisy signals,} Inverse Problems {\bf 24}, 015011 (2008).
%
\bibitem{Beck09} A. Beck and M. Teboulle, {\em A Fast Iterative Shrinkage-Thresholding Algorithm for Linear Inverse Problems,} SIAM J. Img. Sci. {\bf 2}, 183--202 (2009), % DOI 10.1137/080716542.
%
%\bibitem{Bensen07} G. D. Bensen, M. H. Ritzwoller, M. P. Barmin, A. L. Levshin, F. Lin, M. P. Moschetti, N. M. Shapiro, Y. Yang, {\em Processing seismic ambient noise data to obtain reliable broad-band surface wave dispersion measurements,} Geophysical Journal International {\bf 169}, 1239?1260 (2007).% https://doi.org/10.1111/j.1365-246X.2007.03374.x
%
%\bibitem{Blomgren02} P. Blomgren, G. Papanicolaou, Hongkai Zhao, {\em  Super-resolution in time-reversal acoustics,} 
%The Journal of the Acoustical Society of America {\bf 111}, 230 (2002).
%
\bibitem{Borcea02}  L. Borcea,  G. Papanicolaou, C. Tsogka, and James Berryman, {\em Imaging and time reversal in random media}, Inverse Problems {\bf 18}, 1247-1279 (2002). %DOI 10.1088/0266-5611/18/5/303
%
\bibitem{Borcea05} L. Borcea, G. Papanicolaou and C. Tsogka, {\em Interferometric array imaging in clutter}, Inverse Problems {\bf 21}, 1419-1460 (2005) 
%\bibitem{Borcea21} Borcea, Liliana (1-MI); Garnier, Josselin (F-IPP-CMA)
%Imaging in random media by two-point coherent interferometry. (English summary) 
%SIAM J. Imaging Sci. 14 (2021), no. 4, 1635–1668. 
%
\bibitem{Borcea11}L. Borcea, J. Garnier, G. Papanicolaou and C. Tsogka, {\em Enhanced Statistical Stability in Coherent Interferometric Imaging},  Inverse Problems {\bf 27}, 085004 (2011)
%
\bibitem{Borg} I. Borg and P. Groenen, {\em Modern Multidimensional Scaling: Theory and Applications},  Springer Series in Statistics, (2005)
%
\bibitem{Born70}M. Born and E. Wolf, {\em Principles of Optics}, Academic Press, (1970)
%%\bibitem{Cai11}
%%T. T. Cai and L. Wang, {\em Orthogonal matching pursuit for sparse signal recovery with noise,} IEEE Trans. Information Theory {\bf 57}, 4680--4688 (2011).
%
%\bibitem{Chaigne14} T. Chaigne, O. Katz, A. Boccara, et al., {\em Controlling light in scattering media non-invasively using the photoacoustic transmission matrix,} Nature Photon {\bf 8}, 58--64 (2014). %https://doi.org/10.1038/nphoton.2013.307
%
%\bibitem{Chen01} S. S. Chen, D. L. Donoho, and M. A. Saunders, {\em Atomic decomposition by basis pursuit}, SIAM Rev. {\bf 43}, pp. 129--159 (2001).
%
\bibitem{Davis97} G. Davis, S. Mallat, and M. Avellaneda, {\em Adaptive greedy approximations,} Journal of Constructive Approximation {\bf 13}, 57--98 (1997).
%
\bibitem{Daubechies04} I. Daubechies, M. Defrise, and C. De Mol, {\em An iterative thresholding algorithm for linear inverse problems with a sparsity constraint,} Comm. Pure Appl. Math. {\bf 57}, 1413--1457 (2004).
%
%\bibitem{Duvall93} T. L. Duvall, S. M. Jefferies, J. W. Harvey, and M. A. Pomerantz, {\em Time-distance helioseismology}, Nature {\bf 362}, 430--432 (1993).
%
%\bibitem{Efron04} B. Efron, T. Hastie, I. M. Johnstone, and R. Tibshirani, {\em Least angle regression,} The Annals of Statistics {\bf 32}, 407--499 (2004).
%
%%\bibitem{Engan00}
%%K. Engan, S. Kjersti, and J. Husøy, {\em Multi-frame compression: theory and design,} Signal Processing. {\bf 80}, 2121--2140 (2000). 
%
\bibitem{Engan00}  K. Engan, S.O. Aase and J. Hakon Husoy, {\em Method of optimal directions for frame design,} in  1999 IEEE International Conference on Acoustics, Speech, and Signal Processing. Proceedings. ICASSP99 (Cat. No.99CH36258), {\bf 5}, 2443--2446 (1999). 
%
%\bibitem{Fink92}  M. Fink,  {\em Time reversal of ultrasonic field--Part I: Basic principles,} IEEE Trans. on Ultrasonics, ferroelectrics, and frequency control, {\bf 39} (1992), pp.~555--566.
%
%\bibitem{Fink92b} F. Wu, J.L. Thomas and M. Fink, {\em Time reversal of ultrasonic fields. Il. Experimental results}, IEEE Transactions on Ultrasonics Ferroelectrics and Frequency Control,  {\bf 39}, 567--578 (1992)
%
\bibitem{Fink00} M. Fink, D. Cassereau, A. Derode, C. Prada, P. Roux, M. Tanter,
J.-L. Thomas and F. Wu, {\em Time-reversed acoustics}, Rep. Prog. Phys. {\bf 63}, 1933-1995 (2000). 
%
%\bibitem{Garnier09} J. Garnier and G. Papanicolaou, {\em Passive Sensor Imaging Using Cross Correlations of Noisy Signals in a Scattering Medium,} SIAM Journal on Imaging Sciences {\bf 2}, 396--437 (2009).
%
%%\bibitem{Garnier12} J. Garnier and G. Papanicolaou, {\em Correlation-based virtual source imaging in strongly scattering random media,}
%%Inverse Problems {\bf 28},  075002 (2012).
%
%\bibitem{Jaeger15} M. Jaeger, E. Robinson,  H. G. Akarçay, M. Frenz M, {\em  Full correction for spatially distributed speed-of-sound in echo ultrasound based on measuring aberration delays via transmit beam steering,} Phys Med Biol.  {\bf 60}, 4497--4515 (2015). %doi: 10.1088/0031-9155/60/11/4497. Epub 2015 May 19. PMID: 25989072.
%
%\bibitem{Jang13} J. Jang, J. Lim, H. Yu, H. Choi, J. Ha, J.H. Park, W.Y. Oh, W. Jang, S. Lee, Y. Park, {\em Complex wavefront shaping for optimal depth-selective focusing in optical coherence tomography,} Optics Express {\bf 21}  2890--2902 (2013).
%
%\bibitem{Ji10} N. Ji, D.E. Milkie, and E. Betzig, {\em Adaptive optics via pupil segmentation for high-resolution imaging in biological tissues,} Nature Methods {\bf 7}  141--184 (2010).
%
%\bibitem{Kosovichev99}  A. G. Kosovichev, T. L. Duvall, P. H. Scherrer, {\em Time-distance helioseismology,} Advances in Space Research {\bf 24},  163--171 (1999). 
%
%\bibitem{Kruskal} B. Kruskal, {\em Multidimensional scaling by optimizing goodness of fit to a nonmetric hypothesis,} Psychometrika {\bf 29}, 1--27 (1964). 
%
%\bibitem{Lambert20} W. Lambert, L. Cobus, T. Frappart, M. Fink, A. Aubry, {\em Distortion matrix approach for ultrasound imaging of random scattering media,} Proc. Natl. Acad. Sci. U.S.A. {\bf 117}, 14645-56 (2020). % 201921533. 10.1073/pnas.1921533117. 
%
%\bibitem{Lerosey04} G. Lerosey, J. de Rosny, A. Tourin, A. Derode, G. Montaldo, and M. Fink, {\em Time Reversal of Electromagnetic Waves}, Rev. Lett. {\bf 92}, 193904 (2004).
%
%\bibitem{Masoy05} S.E. Måsøy, T. Varslot, B. Angelsen, {\em  Iteration of transmit-beam aberration correction in medical ultrasound imaging,} J Acoust Soc Am.  Jan; {\bf 17}, 450-461 (2005). %doi: 10.1121/1.1823213. PMID: 15704438.
%
%%\bibitem{Mallat93}
%%S. Mallat and Z. Zhang, {\em Matching pursuits with time-frequency dictionaries,} IEEE Trans. Signal Process {\bf 41}, 3397--3415 (1993).
%
\bibitem{Moscoso12} M. Moscoso, A. Novikov, G. Papanicolaou, and L. Ryzhik, {\em A differential equations approach to l1-minimization with applications to array imaging,} Inverse Problems {\bf 28}, 105001 (2012).
%
\bibitem{Moscoso17} M. Moscoso, A. Novikov, G. Papanicolaou, and C. Tsogka, {\em Multifrequency Interferometric Imaging with Intensity-Only Measurements,} SIAM J. Img. Sci.  {\bf 10}, 1005--1032 (2017).
%
%\bibitem{Muller74} R. A. Muller and A. Buffington, {\em Real-time correction of atmospherically degraded telescope images through image sharpening,} J. Opt. Soc. Am. {\bf 64}, 1200--1210 (1974).
%
%\bibitem{Webb} J. A. Newman, Q. Luo, and K. J. Webb, “Imaging hidden objects with spatial speckle intensity correlations over object position,” Phys. Rev. Lett. 116, 73 902 (2016).
%
\bibitem{Novikov23}A. Novikov, S. White, {\em Spectral subspace dictionary learning,} Proceedings of Machine Learning Research,   34th International Conference on Algorithmic Learning Theory, 1--36 (2023).
%
%%%\bibitem{Hampson21} K. M. Hampson, R. Turcotte, D. T. Miller, et al,  {\em Adaptive optics for high-resolution imaging,} Nat Rev Methods Primers {\bf 1} (2021). %https://doi.org/10.1038/s43586-021-00066-7
%
\bibitem{Oh10} S. Oh, A. Montanari and A. Karbasi, {\em Sensor network localization from local connectivity: Performance analysis for the MDS-MAP algorithm,} 2010 IEEE Information Theory Workshop on Information Theory (ITW 2010, Cairo), Cairo, Egypt, pp. 1--5 (2010).
%
\bibitem{Osborne00} M. R. Osborne, B. Presnell, and B. A. Turlach, {\em On the lasso and its dual,} Journal of Computational and Graphical Statistics {\em 9}, 319--337 (2000).
%
%\bibitem{Popoff10} S. M. Popoff, G. Lerosey, R. Carminati, M. Fink, A. C. Boccara, S. Gigan, {\em Measuring the transmission matrix in optics: an approach to the study and control of light propagation in disordered media,} Phys. Rev. Lett. {\bf 104}, 100601 (2010).
%
%\bibitem{Rickett99} J. Rickett and J. Claerbout, {\em Acoustic daylight imaging via spectral factorization: Helio- seismology and reservoir monitoring,} The Leading Edge {\bf 18}, 957--960 (1999).
%
%\bibitem{Roux05} P. Roux, K. G. Sabra, and W. A. Kuperman, {\em Ambient noise cross correlation in free space: Theoretical approach,} The Journal of the Acoustical Society of America {\bf 117}, 79--84 (2005).
%
%
%\bibitem{Schuster04} G. T. Schuster, J. Yu, J. Sheng, and J. Rickett, {\em Interferometric daylight seismic imaging,} Geophysical Journal International {\bf 157}, 832--852 (2004).
%
\bibitem{Shang03} Y. Shang, W. Ruml, Y. Zhang, and M. P. J. Fromherz, {\em Localization from mere connectivity,} in MobiHoc `03: Proceedings of the 4th ACM international symposium on Mobile ad hoc networking \& computing. New York, NY, USA: ACM, pp. 201--212 (2003).
%
%\bibitem{Spielman12} D. A. Spielman, H. Wang, and J. Wright, {\em Exact recovery of sparsely-used dictionaries,} In Proc. of Conf. on Learning Theory (2012).
%
%\bibitem{Staley13} J. Staley, E. Hondebrink, W. Peterson, and W. Steenbergen, {\em Photoacoustic guided ultrasound wavefront shaping for targeted acousto-optic imaging,} Opt. Express {\bf 21}, 30553--30562 (2013).
%
\bibitem{Stansfield47} R.G. Stansfield, Statistical theory of d.f. fixing, Journal of the Institution of Electrical Engineers {\bf 94}, 762--770 (1947).
%
%\bibitem{Tyson}R. K. Tyson, Principles of Adaptive Optics, CRC Press, (2015).
%
%\bibitem{Vellekoop07} I. M. Vellekoop, and A. P.  Mosk,  {\em Focusing coherent light through opaque strongly scattering media,} Opt. Lett. {\bf 32}, 2309--2311 (2007).
%
%%%\bibitem{Yaqoob08} Z. Yaqoob and D. Psaltis, {\em Optical phase conjugation for turbidity suppression in biological samples,} Nat Photonics {\bf 2}, 110--115 (2008).
%%
%%%\bibitem{Katz12} O. Katz, E. Small, Y. Silberberg, {\em Looking around corners and through thin turbid layers in real time
%%%with scattered incoherent light,} Nat. Photonics {\bf 6}, 549--553 (2012).
%
%\bibitem{Weaver01} R. L. Weaver, and O. I. Lobkis, {\em Ultrasonics without a source: thermal fluctuation correlations at MHz frequencies,} Physical Review Letters 87, 134301-1--134301-4 (2001).
%
\bibitem{Wu19} P. Wu, S. Su, Z. Zuo, X. Guo, B. Sun and X. Wen, Time Difference of Arrival (TDoA) Localization Combining Weighted Least Squares and Firefly Algorithm, Sensors {\bf 19}, 2554 (2019).
%
\bibitem{Yeredor11} A. Yeredor and E. Angel, Joint TDOA and FDOA Estimation: A Conditional Bound and Its Use for Optimally Weighted Localization,IEEE Transactions on Signal Processing {\bf 59}, 1612--1623 (2011). %doi: 10.1109/TSP.2010.2103069.
%
%bibitem{Yu14}  H. Yu, J. Jang, J. Lim, J.-H. Park, W. Jang, J.-Y. Kim, Y. Park, {\em Depth-enhanced 2-D optical coherence tomography using complex wavefront shaping,} Optics Express {\bf 22}  7514--7523 (2014).
%
%%%\bibitem{Fink97} M. Fink, {\em Time Reversed Acoustics,} Physics Today {\bf 50}, 34--40 (1997)
%%
%%
%%%\bibitem{Barak15}
%%%B. Barak, J. A. Kelner, and D. Steurer, {\em Dictionary learning and tensor decomposition via the sum-of-squares method,} Proceedings of the Forty-Seventh Annual ACM Symposium on Theory of Computing, STOC ’15, 143--151 (2015). 






\end{thebibliography}
\end{document}